%% file: acl_latex.tex
\documentclass[11pt]{article}

\PassOptionsToPackage{table}{xcolor}
\usepackage[preprint]{acl}

\usepackage{times}
\usepackage{latexsym}

\usepackage[T1]{fontenc}
\usepackage[utf8]{inputenc}

\usepackage{microtype}

\usepackage{inconsolata}
\usepackage{fvextra}

\usepackage{graphicx}

\usepackage{booktabs}
\usepackage{multirow}

\DefineVerbatimEnvironment{PromptBox}{Verbatim}{
  breaklines=true,
  breakanywhere=true,
  fontsize=\footnotesize,
  frame=single,
  framesep=2mm,
  commandchars=\\\{\}
}
\DefineVerbatimEnvironment{PromptFullBox}{Verbatim}{
  breaklines=true,
  breakanywhere=true,
  fontsize=\scriptsize,
  frame=single,
  framesep=1mm
}

\makeatletter
\newif\if@appendixtoc
\@appendixtocfalse
\newcommand{\appendixTOCstart}{\global\@appendixtoctrue}

\let\old@contentsline\contentsline
\renewcommand{\contentsline}[4]{%
  \if@appendixtoc
    \old@contentsline{#1}{#2}{#3}{#4}%
  \fi
}
\makeatother

\title{Large Language Models Do Not Always Need Readable Language}

\author{
  \textbf{Jiayi Zhu}\textsuperscript{1}
  \quad
  \textbf{Haoxuan Peng}\textsuperscript{2}
  \quad
  \textbf{Junxi Wang}\textsuperscript{3}
  \quad
  \textbf{Liang Ke}\textsuperscript{4}
  \quad
  \textbf{Chen Zhang}\textsuperscript{5}
  \quad
  \textbf{Linfeng Zhang}\textsuperscript{1}\thanks{\ \ Corresponding author.}
  \\
  {\small
  \textsuperscript{1}Shanghai Jiao Tong University;
  \textsuperscript{2}The University of Sydney;
  \textsuperscript{3}Hefei University of Technology} \\
  {\small
  \textsuperscript{4}Xi'an Jiaotong University;
  \textsuperscript{5}Nanjing University} \\
  \texttt{zhanglinfeng@sjtu.edu.cn}
}

\begin{document}
\maketitle

\begin{abstract}
Large language models (LLMs) are commonly prompted and interfaced with human-readable natural language, even when the intended reader is another model. This paper investigates whether semantic information can be encoded in compact, non-standard textual forms that sacrifice human readability while remaining recoverable by LLMs. We refer to this class of model-centric textual representations as \textbf{BabelTele}, approached here not as a fixed protocol but as an empirical probe into LLMs' capacity to generate and interpret such representations. Through readability diagnostics, model likelihood measures, human questionnaires, and downstream task evaluations, we find that BabelTele can substantially depart from ordinary natural language while preserving core semantics for instruction-tuned LLMs. As a task-agnostic representational paradigm, BabelTele demonstrates high information density, maintaining 99.5\% semantic fidelity even when the text volume is condensed to 27.9\% of its original length. We further evaluate its semantic robustness in cross-model transfer, agent memory, and multi-agent communication. Results suggest that BabelTele can reduce context overhead while generally maintaining reliable downstream performance, although its effectiveness depends on the compressor-reader pair and task setting. These findings indicate that human readability, natural-language typicality, and model-side semantic recoverability can be partially decoupled, opening a path toward model-native representations in future exploration of LLM systems.
\end{abstract}

\section{Introduction}

Large language models (LLMs) have become a dominant interface in contemporary intelligent systems. Since GPT-3 demonstrated strong few-shot generalization through text-based prompting \citep{brown2020fewshot}, the field has largely followed a unified paradigm: knowledge is represented in natural language, instructions are issued in natural language, and model outputs are returned in natural language. Subsequent alignment methods and dialogue-oriented models further reinforced this design, optimizing model behavior toward controllability, and readability \citep{ouyang2022instructions, touvron2023llama2}.

% Human language is not merely an information carrier, but a medium evolved to accommodate perception, memory, and social coordination. Consequently, natural language contains substantial redundancy. Many words and syntactic structures add little direct factual information; instead, they help readers follow, remember, and disambiguate content. From this perspective, the very features that make language easy for people to use can also make it less compact as an information channel.

\begin{figure}[t]
  \includegraphics[width=\columnwidth]{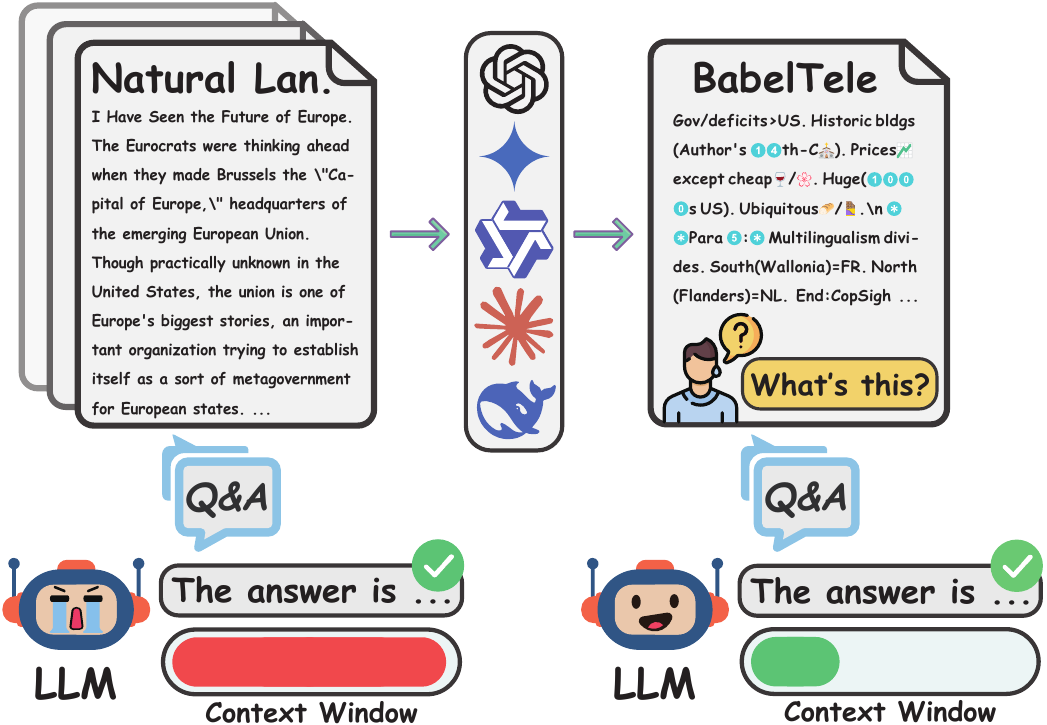}
  \caption{As illustrated, \textbf{BabelTele} representation differs substantially from verbose natural language: the text is significantly more compact, indicating a much higher information density. While the compressed representation is much less human-readable, it remains well-interpreted by LLMs, which can understand the original meaning without any distortion.}
  \label{fig1}
  \vspace{-15pt}
\end{figure}

However, natural language optimized for human communication is not necessarily an efficient representation for model processing. Human language contains substantial redundancy: complete syntax, discourse markers, and narrative coherence all help people follow, remember, and disambiguate information. These properties are valuable for human readers, but they reduce semantic density. From an information-theoretic view, communication systems aim to transmit information efficiently under channel constraints \citep{shannon1948comm}. More recent work also connects language modeling with compression, suggesting that strong language models can capture compact statistical structure in data \citep{deletang2024language}. This raises a central question: if the receiver is an LLM rather than a human, must semantic information still be encoded in fully human-readable natural language?

% However, if the primary receiver is a Large Language Model (LLM) rather than a human, the value of these redundancies changes. Complete syntax, coherent narrative flows, and repetitive semantic cues are not strictly necessary for an LLM to recover semantics. For models, information could potentially be encoded into a highly compact, non-human-readable form that still preserves essential semantic structures.

This question becomes especially relevant in long-context and agentic systems, where context overhead is a persistent bottleneck. LLMs are increasingly deployed to process lengthy documents, maintain memory, and exchange intermediate states in multi-agent workflows, yet models do not always use long contexts robustly \citep{liu2024lost}, and agent systems further intensify this pressure through natural-language memory \citep{park2023generative}, context management \citep{packer2023memgpt}, and inter-agent message passing \citep{wu2024autogen}. The idea of sacrificing readability for efficiency, however, is not new: telegraphic language, mathematical notation, and code all demonstrate that fluent prose is only one possible surface form for communication. This suggests a broader possibility that intermediate representations in LLM systems may be optimized not for human readability, but for model decodability and semantic density.

% This issue is particularly relevant in long-context scenarios. LLMs are increasingly used to process lengthy documents, maintain persistent memory, invoke tools, and exchange intermediate states in multi-agent systems. Currently, this information is stored and transmitted in standard natural language, continually consuming limited context windows. As task lengths and system complexity grow, human-readable redundancy transforms from a linguistic feature into a significant computational cost.

% This is not the first time communication has traded readability for efficiency. When communication costs were high, people developed condensed formats. Telegraphs omitted grammar in favor of keywords, order, and shared context; mathematical symbols, abbreviations, code, and tables further compressed complex relationships into minimal units. Evidently, fluent natural sentences are not the only valid form of language.

% Existing research has explored context compression through token selection, prompt rewriting, retrieval, summarization, and learned compact representations. While these works confirm the redundancy in natural language, they typically focus on \emph{what} information to retain or how to encode context into continuous internal representations. BabelTele asks a different question: can retained semantics move beyond natural-language form and be re-encoded into a leaner, hyper-dense \emph{textual} representation that remains decodable by LLMs?

Existing context-compression methods provide important precedents. Prior work has explored reducing prompt length by removing redundant spans, rewriting retrieved content, or selecting informative tokens, demonstrating that natural-language inputs contain substantial compressible redundancy \citep{li2023compressing, jiang2023llmlingua, xu2024recomp}. Nevertheless, most such methods still operate within the conventions of human-readable natural language. A separate line of work explores activation-based or learned internal representations, but these approaches often require additional training, special tokens, or access to hidden states, limiting their applicability in black-box API settings and heterogeneous model ecosystems.

% Motivated by this question, this paper introduces and investigates \textbf{BabelTele}, a class of model-centric rather than human-centric textual representations. BabelTele uses abbreviations, symbols, cross-lingual snippets, structured tags, and non-standard syntax. To humans, it appears disjointed and unnatural; yet, for advanced LLMs, its entities, relations, events, and contextual clues remain recoverable for downstream tasks.

% In our experiments, we instantiate BabelTele using a specifically constrained prompt. The instruction asks the model to preserve core semantics while allowing the surface form to deviate from human readability, producing a compact, symbolic, and model-oriented representation. This serves as a practical entry point for observing the BabelTele phenomenon (full prompt in Appendix).

% This instantiation should not be confused with the concept itself. BabelTele is not defined by a single prompt; the prompt acts only as an empirical probe that materializes the phenomenon. Our focus is not on a specific ''prompt trick'', but on the existence, properties, and practical benefits of representations that sacrifice human readability for model-recoverable density.

Motivated by this gap, we investigate \textbf{BabelTele}, a class of model-centric rather than human-centric textual representations. BabelTele explicitly relaxes human readability as a default constraint, instead encouraging models to encode semantics into compact, non-standard textual forms, potentially combining abbreviations, symbols, cross-lingual fragments, and non-standard syntactic structures. While these forms exhibit low human readability, advanced LLMs can still recover their core semantics for downstream tasks. \textbf{We thus present BabelTele not as a competitive compression method, but as an empirical probe into model-native textual communication}, evaluating it across multiple dimensions including compression ratio, semantic fidelity, human readability, cross-model transferability, and downstream utility in document QA, agent memory compression, and multi-agent communication. Our key takeaways are as follows:
\begin{itemize}
    \item BabelTele emerges as a prompt-accessible phenomenon: by removing human readability as a default constraint, LLMs spontaneously produce opaque but semantically dense textual representations under a black-box interface.

    \item BabelTele exhibits robust cross-model transferability across diverse proprietary and open-weight model families in a zero-shot manner. Representations compressed by one model can be reliably interpreted by another without any fine-tuning or model-specific adaptation, suggesting that BabelTele captures a form of semantic encoding that generalizes across heterogeneous architectures.

    % \item BabelTele proves effective in practical long-context scenarios, including document QA, agent memory compression, and multi-agent communication, suggesting it as a promising direction beyond existing natural-language prompt compression methods.

    \item BabelTele retains semantics in document QA, agent memory, and multi-agent communication, demonstrating that current LLMs can process highly dense text without relying on human readability.
\end{itemize}

\section{Related Works}

% To produce a PDF file, pdf\LaTeX{} is strongly recommended (over original \LaTeX{} plus dvips+ps2pdf or dvipdf).
% The style file \texttt{acl.sty} can also be used with
% lua\LaTeX{} and
% Xe\LaTeX{}, which are especially suitable for text in non-Latin scripts.
% The file \texttt{acl\_lualatex.tex} in this repository provides
% an example of how to use \texttt{acl.sty} with either
% lua\LaTeX{} or
% Xe\LaTeX{}.

\subsection{Prompt and Context Compression}
%Prompt compression has been widely studied as a way to reduce inference cost and improve the effective use of limited context windows. Hard compression methods usually keep the compressed prompt as discrete text. Selective Context () filters tokens or sentences according to self-information. LLMLingua () performs coarse-to-fine token-level prompt compression. LongLLMLingua () extends this idea to long-context scenarios by increasing the density and placement of key information. LLMLingua-2 () formulates prompt compression as a token classification problem and improves task-agnostic faithfulness through data distillation. Instruction-aware contextual compression further conditions compression on downstream instructions (). These approaches mainly reduce prompts by deleting or selecting lexical units from the original text.

%Other work studies abstractive or natural-language prompt compression. RECOMP () compresses retrieved documents into extractive or abstractive summaries before feeding them to retrieval-augmented language models. Nano-Capsulator () learns to compress prompts into shorter natural-language capsule prompts. Adaptive context compression methods dynamically choose compression rates for RAG inputs according to query complexity or evidentiality (). Unlike these methods, BabelTele does not aim to preserve natural-language readability. It instead asks whether LLMs can generate and consume compact semantic strings that are opaque to humans but still interpretable by LLMs.

Prompt compression has been widely studied as a way to reduce inference cost and improve the effective use of limited context windows \citep{li2023compressing, jiang2023llmlingua, jiang2024longllmlingua, pan2024llmlingua2, hou2024instructionaware}.
% 这个后面写下面的五个引用，直接引用即可，然后例子举出两个即可，不用全写出来
Hard compression methods usually keep the compressed prompt as discrete text. Selective Context \citep{li2023compressing} filters tokens or sentences according to self-information. LLMLingua \citep{jiang2023llmlingua} performs coarse-to-fine token-level prompt compression.
%LongLLMLingua () extends this idea to long-context scenarios by increasing the density and placement of key information. LLMLingua-2 () formulates prompt compression as a token classification problem and improves task-agnostic faithfulness through data distillation. Instruction-aware contextual compression further conditions compression on downstream instructions (). 
These approaches mainly reduce prompts by deleting or selecting lexical units from the original text.

Other work studies abstractive or natural-language prompt compression \citep{xu2024recomp, chuang2024nanocap, zhang2024adacomp, jeong2025ecorag, guo2025enhanrag}.
RECOMP \citep{xu2024recomp} compresses retrieved documents into summaries, while Nano-Capsulator \citep{chuang2024nanocap} learns shorter natural-language capsule prompts.
% 下面这个太长了，浓缩到我上面的即可
%RECOMP () compresses retrieved documents into extractive or abstractive summaries before feeding them to retrieval-augmented language models. Nano-Capsulator () learns to compress prompts into shorter natural-language capsule prompts. Adaptive context compression methods dynamically choose compression rates for RAG inputs according to query complexity or evidentiality (). 
Unlike these methods, BabelTele does not aim to preserve natural-language readability. It instead asks whether LLMs can generate and consume compact semantic strings that are opaque to humans but still interpretable by LLMs.

We also consider related work on learned and latent context compression, retrieval and memory systems, symbolic prompting and LLM-native communication. Due to space limitations, they are provided in Appendix~\ref{App_relatedworks}.

\section{Methodology: Eliciting LLM-Native Representations}

\subsection{Relaxing the Readability Prior}
Given an input document $x$, conventional text compression employs a model $C$ to produce a shorter sequence $z$ that preserves task-relevant semantics for a reader model $R$. Most existing methods implicitly encourage $z$ to remain close to the natural language distribution, keeping the compressed text reasonably readable to humans.

BabelTele formulates compression as a \textbf{readability-relaxed semantic projection}. When $R$ is an LLM rather than a human reader, we hypothesize that the human-readability prior can be relaxed. Instead of optimizing for fluency or natural-language typicality, BabelTele prioritizes information density and model-side semantic recoverability. While $z$ remains discrete text, its surface form can be deviated from conventional prose. Furthermore, as different LLMs are trained on overlapping linguistic, symbolic and factual structures, such representations may exhibit partial cross-model decodability across model families.

\subsection{Principles of Symbolic Collapse}
Rather than treating BabelTele as a singular, manually engineered ``prompt trick,'' we define it as a family of high-density representations induced by relaxing linguistic constraints. To materialize this phenomenon in a black-box setting, we design instructional probes that encourage the compressor $C$ to produce model-readable encodings based on the following principles:

\textbf{Omnilingual Lexical Selection:} Relaxing single-language constraints and selecting high-density lexical units across languages and scripts.

\textbf{Symbolic Collapse:} Replacing verbose linguistic structures with compact symbols, emojis, mathematical/logical operators, and punctuation.

\textbf{Recoverable Semantic Density:} Preserving recoverable semantic details so that capable LLMs can interpret the compressed output without an external codebook.

By applying these constraints through zero-shot prompting, we encourage LLMs to externalize semantic information into a compact surface form more effectively. This approach requires no gradient updates or tokenizer modifications, allowing us to investigate the existence and utility of LLM-native textual representations across heterogeneous models. To illustrate this collapse, consider the following micro-example:

\noindent\includegraphics[width=\columnwidth]{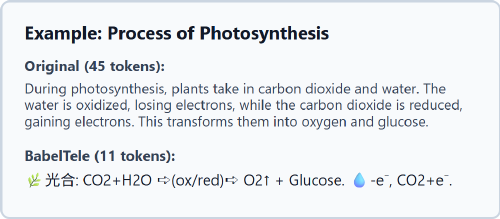}
As shown above, BabelTele relaxes ordinary human syntax and uses multilingual cues, emojis, and relational arrows to form a compact, model-readable semantic graph more densely. Appendix~\ref{App_examples} provides a qualitative example with a source excerpt and the complete BabelTele output.

% \subsection{Footnotes}

% Footnotes are inserted with the \verb|\footnote| command.\footnote{This is a footnote.}
% \begin{table*}[t]
% \fontsize{8}{13}\selectfont
% \setlength{\tabcolsep}{7pt}
% \centering
% \vspace{3pt} 
% \begin{tabular}{lcccc}
%   \toprule[1.5pt]
%   \textbf{Method}
%        & \textbf{Token count (k) ({\color{green!40!black}$\downarrow$})}
%        & \textbf{Retention ratio ({\color{green!40!black}$\downarrow$})}
%        & \textbf{Acc. ({\color{green!40!black}$\uparrow$})}
%        & \textbf{\% ({\color{green!40!black}$\uparrow$})} \\
%   \midrule
%   \rowcolor{gray!15}
%   Original
%         &1705.0&100\%&64.86&100.00\% \\
%   LLMLingua2
%         &\textbf{49.8}&\textbf{3.0}\%&54.05&72.90\% \\
%   \rowcolor{green!3}
%   BabelTele
%         &65.1&3.4\%&\textbf{63.56}&\textbf{96.74}\% \\
%   \bottomrule[1.5pt]
% \end{tabular}
% \caption{\textbf{Performance comparison of different methods on the Short subset of Single-Document QA in the LongBench v2 benchmark.} \textbf{}\% represents the performance retention rate relative to the original result. The best result is highlighted in \textbf{bold black font}.}
% \label{table1}
% \vspace{-10pt} 
% \end{table*}

\section{Experiments}

\subsection{Experimental Setup}

We evaluate BabelTele across multiple long-context benchmarks under a task-agnostic protocol, where the compressor processes source documents without access to downstream questions. We compare against standard baselines and report token retention, QA accuracy, and reader chain-of-thought token overhead across diverse evaluator model families. Full details are in Appendix~\ref{App_setup}.

% We evaluate BabelTele on QuALITY, LongBench v2, LoCoMo, DeepResearch, and MeetingBank. Crucially, we employ a \textbf{task-agnostic protocol}: the compressor (Gemini 3.1 Pro) processes the source document without seeing downstream questions. We compare against the original text, natural-language summaries, and LLMLingua-2~\citep{pan2024llmlingua2}. We report token retention, downstream QA accuracy, and reader chain-of-thought token overhead. Evaluator models span diverse families (e.g., GPT, Claude, Qwen, DeepSeek, Kimi). Full implementation details are in Appendix~\ref{App_setup}.

\subsection{Symbolic Collapse: Separating Human Readability from Model Decodability}

In our early observations, we found that although BabelTele is usually difficult for humans to read directly, LLMs can still use it to answer questions, recover details, and perform a certain degree of reasoning. Therefore, this section mainly examines whether human readability, natural-language distribution typicality, and model semantic recoverability can be decoupled, rather than focusing on the compression efficiency of BabelTele itself.

We select 10 long-text samples from the QuALITY \citep{pang2022quality} dataset, each containing 3 multiple-choice QA items, and construct three input formats for each sample: the original text, a natural-language summary, and a BabelTele representation generated using the default compression prompt in Appendix~\ref{App_prompt_default}.

\begin{table}[h]
\fontsize{7.5}{9}\selectfont
\setlength{\tabcolsep}{5pt}
\renewcommand{\arraystretch}{1.0}
\centering
\vspace{3pt}
\newcommand{\pplcellpos}[2]{#1 {\scriptsize(\color{green!40!black}#2)}}
\newcommand{\pplcellneg}[2]{#1 {\scriptsize(\color{red!70!black}#2)}}
\begin{tabular}{lccc}
  \toprule[1.5pt]
  \textbf{Model} & \textbf{Orig}. & \textbf{Summ}. & \textbf{BabelTele} \\
  \midrule
  Llama-3-8B & 9.63 & \cellcolor{gray!15}\pplcellpos{11.32}{+17.5}\% & \cellcolor{green!3}\pplcellpos{176.60}{+1{,}733.9}\% \\
  Qwen2-7B & 10.93 & \cellcolor{gray!15}\pplcellpos{12.44}{+13.8}\% & \cellcolor{green!3}\pplcellpos{236.23}{+2{,}061.1}\% \\
  Qwen2.5-7B & 10.43 & \cellcolor{gray!15}\pplcellpos{11.56}{+10.8}\% & \cellcolor{green!3}\pplcellpos{209.30}{+1{,}906.7}\% \\
  Qwen3-8B & 14.94 & \cellcolor{gray!15}\pplcellpos{15.22}{+1.9}\% & \cellcolor{green!3}\pplcellpos{301.22}{+1{,}916.2}\% \\
  Qwen3-32B & 11.80 & \cellcolor{gray!15}\pplcellneg{11.07}{-6.2}\% & \cellcolor{green!3}\pplcellpos{220.76}{+1{,}770.8}\% \\
  Kimi K2P6 & 6.93 & \cellcolor{gray!15}\pplcellpos{7.76}{+11.9}\% & \cellcolor{green!3}\pplcellpos{127.96}{+1{,}746.5}\% \\
  DeepSeek V4 Pro & 7.67 & \cellcolor{gray!15}\pplcellpos{7.75}{+1.1}\% & \cellcolor{green!3}\pplcellpos{108.95}{+1{,}320.5}\% \\
  GLM 5.1 & 7.89 & \cellcolor{gray!15}\pplcellpos{8.40}{+6.5}\% & \cellcolor{green!3}\pplcellpos{148.23}{+1{,}778.7}\% \\
  \bottomrule[1.5pt]
\end{tabular}
\caption{\textbf{PPL diagnostics across representative base models}. Parentheses indicate the relative change from the original-text PPL within the same model. {\color{green!40!black}Green} denotes an increase. {\color{red!60!black}Red} denotes a decrease.}
\label{tab:symbolic-collapse-distribution}
\vspace{-10pt}
\end{table}

Readability diagnostics in Appendix~\ref{App_details_symbolic} show that BabelTele has much lower surface readability than both original passages and natural-language summaries, with a Dale-Chall score of \textbf{16.70} and a difficult-word ratio of \textbf{80.19}\%. Table~\ref{tab:symbolic-collapse-distribution} further shows that BabelTele is highly unlikely under multiple base language models, yielding order-of-magnitude higher PPL than both original and summary texts. Together, these results indicate that BabelTele is not merely a natural-language summary or shorthand, but a surface representation that substantially departs from conventional English prose and the ordinary natural-language distribution.

\begin{figure}[h]
  \centering
  \includegraphics[width=\columnwidth]{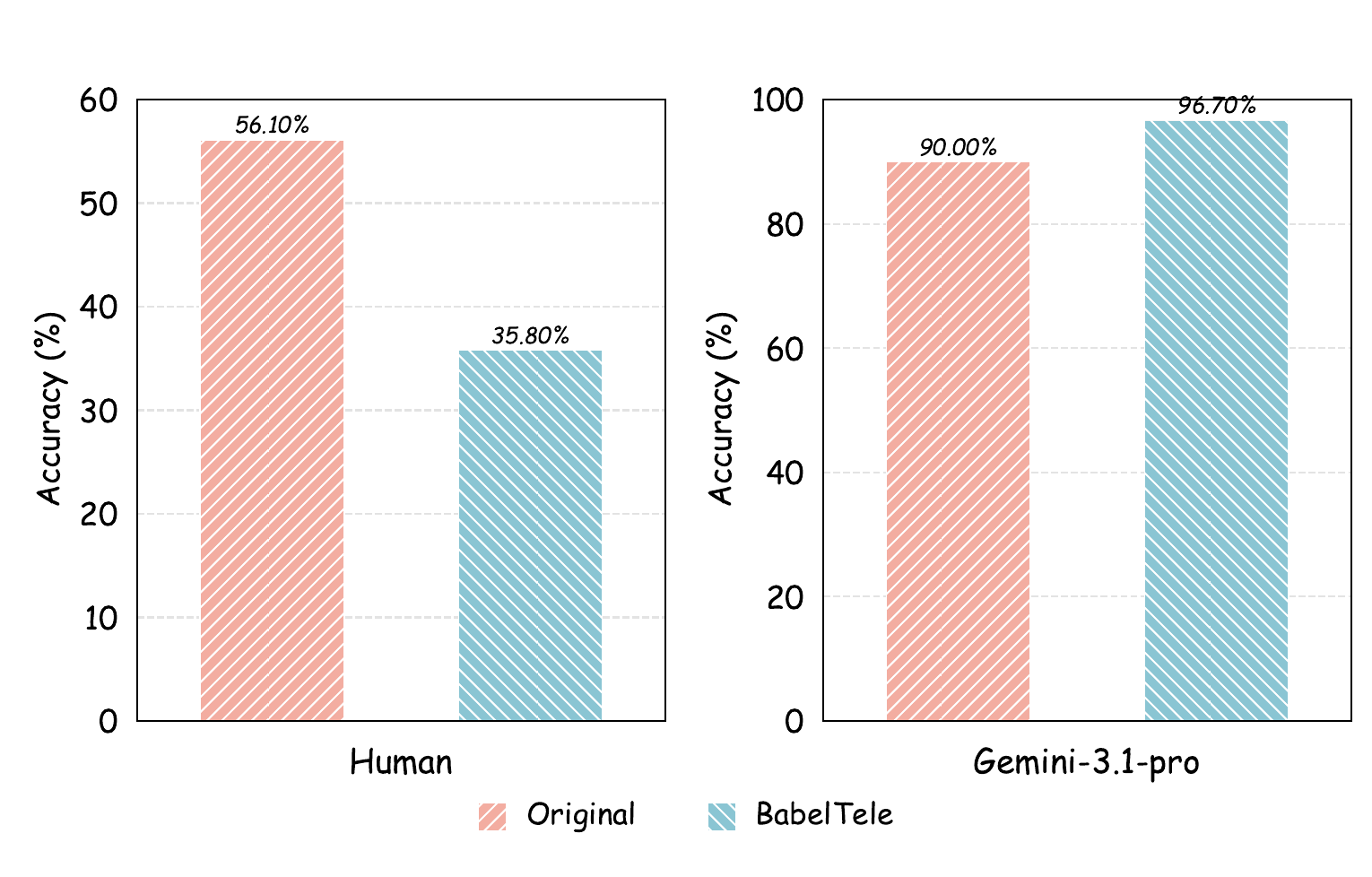}
  \caption{\textbf{QA accuracy for human readers and Gemini 3.1 Pro on original and BabelTele inputs}. The y-axis starts at 25\%, the random-choice baseline for four-option QA. Brackets indicate absolute changes in percentage points.}
  \label{fig:symbolic-collapse-qa-accuracy}
  \vspace{-10pt}
\end{figure}

However, Figure~\ref{fig:symbolic-collapse-qa-accuracy} shows that low readability and low natural-language likelihood do not imply semantic unrecoverability. The human accuracy results were collected through paid questionnaires distributed to university students. Human readers show a QA accuracy drop on BabelTele inputs, whereas Gemini 3.1 Pro \citep{google2026gemini3-1} maintains high accuracy. This suggests that BabelTele is not meaningless gibberish; rather, while sacrificing human readability and natural-language typicality, it still preserves semantic structures that can be decoded and used by instruction-tuned LLMs. More detailed analyses of readability, evaluation metrics, and questionnaire settings are provided in Appendix~\ref{App_details_symbolic}.

\subsection{Efficiency and Cognitive Overhead in Model-Native Compression}
\label{section4_3}

\begin{figure*}[t]
    \centering
    \includegraphics[width=\textwidth]{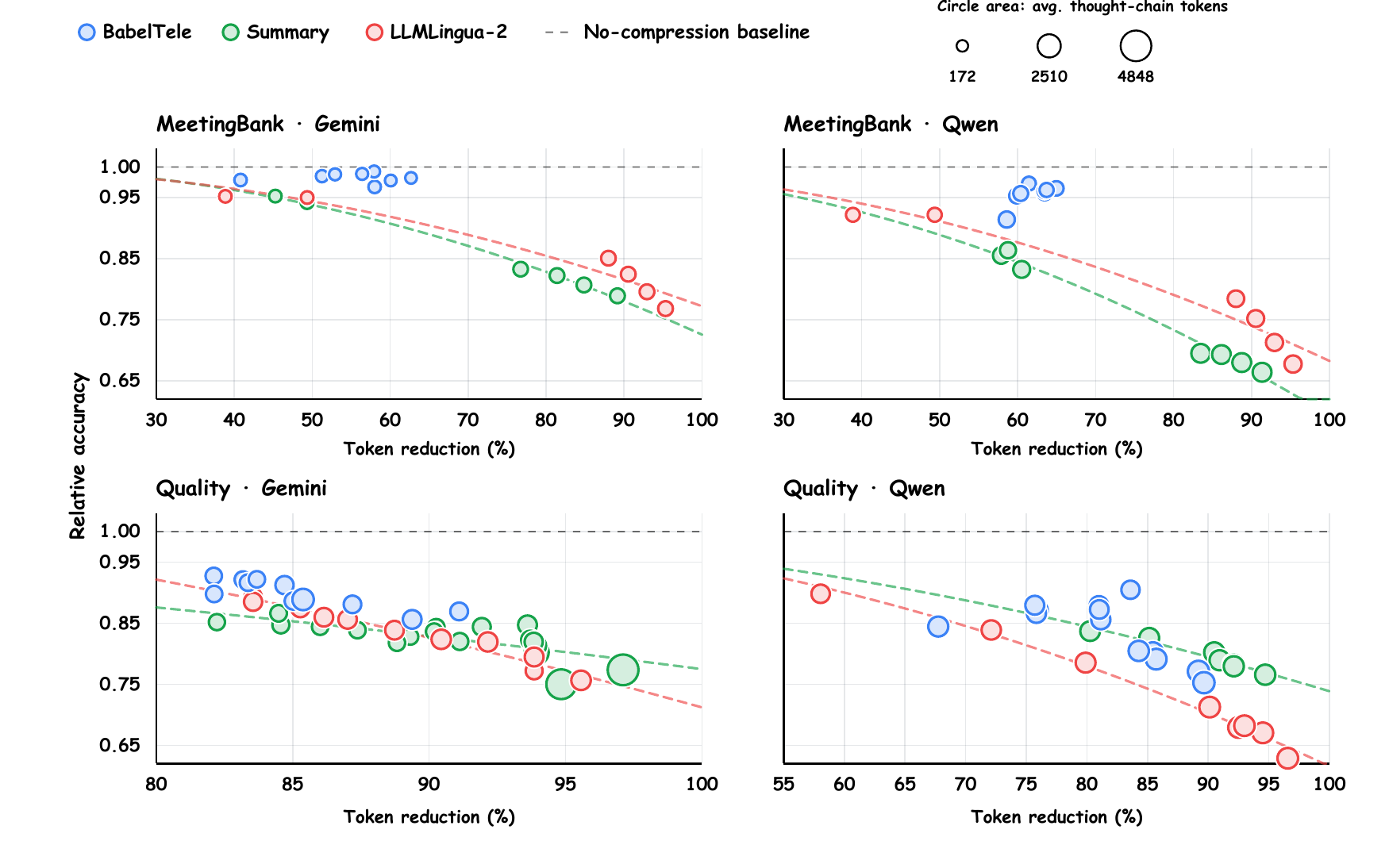}
    \caption{\textbf{Accuracy-retention comparison with response chain-of-thought token scale}. Each panel corresponds to one benchmark-reader setting. Token reduction is computed as one minus the realized context retention ratio, relative accuracy is normalized by the corresponding no-compression baseline, and circle area denotes the average number of response thought-chain tokens. Dashed colored curves indicate method-level fitted trends.}
    \label{fig:compression-scatter-trend}
    \vspace{-10pt}
\end{figure*}

We focus on two questions. First, under different context retention ratios, how much downstream QA accuracy can BabelTele preserve compared with natural-language summaries and LLMLingua-2 \citep{pan2024llmlingua2}? Second, does compression increase the response chain-of-thought tokens used by the reader model? Together, this section evaluates both input-side token savings and answer-side cognitive overhead.

\paragraph{Experimental Setup.}
We evaluate BabelTele on 2,128 QuALITY questions and 2,586 MeetingBank \citep{hu2023meetingbank} multiple-choice questions, conducting 116 experimental runs across the two datasets. We contrast simply prompt-elicited BabelTele representations with standard abstractive summaries and a carefully engineered extractive baseline (LLMLingua-2) under accuracy-retention curves rather than a single compression point, since compression ratios vary across generative methods. For BabelTele and summary baselines, we use the same-model setting, For BabelTele and summary baselines, we use a same-model setting, where each model reads its own compressed output.

BabelTele is evaluated with multiple BabelTele-like prompt variants, so the experiment tests a family of model-readable high-density representations rather than a prompt artifact; additional sweep and prompt details are provided in Appendix~\ref{App_details_native}.

%\paragraph{The Accuracy-Retention Frontier.}
%Figure~\ref{fig:compression-scatter-trend} summarizes the accuracy-retention relation across QuALITY and MeetingBank. Overall, BabelTele forms a more favorable frontier in multiple benchmark-reader settings: it maintains relatively high downstream accuracy under stronger compression, whereas summary and LLMLingua-2 show sharper accuracy degradation as compression becomes more aggressive.

%This trend is especially clear on MeetingBank. For both Gemini and Qwen readers, BabelTele preserves near-baseline performance even under large token reductions. On QuALITY, the gap is more moderate, but BabelTele still maintains competitive relative accuracy across the evaluated compression range. This suggests that the benefit of BabelTele is not limited to a single benchmark format; it can preserve task-relevant information in both meeting-style records and long-document QA.

%The trend induced by multiple prompt variants further supports the family-level interpretation of BabelTele. Rather than relying on one optimized prompt, different BabelTele-like prompts collectively trace a strong accuracy-retention frontier, suggesting that BabelTele is a class of model-readable compressed representations rather than a single prompt trick.

\paragraph{The Accuracy-Retention Frontier}.
Figure~\ref{fig:compression-scatter-trend} summarizes the relation between accuracy and context retention on QuALITY and MeetingBank. Overall, BabelTele forms a more favorable accuracy-retention frontier across multiple benchmarks and reader models. We summarize the following three points:

\textbf{(i) BabelTele maintains higher accuracy under strong compression}. As compression intensifies, summary and LLMLingua-2 show sharper accuracy degradation, whereas BabelTele maintains relatively high downstream QA accuracy. This suggests that BabelTele can better preserve task-relevant semantics while reducing input tokens.

\textbf{(ii) BabelTele's robustness is more evident on MeetingBank}. For both Gemini 3.1 Pro and Qwen-3.5-Plus readers, BabelTele preserves near-original performance even with substantial token reduction. On QuALITY, the advantage is more moderate, but BabelTele still maintains stable relative accuracy across the evaluated compression range. This indicates that the benefit of BabelTele is not limited to a single task format, but applies to both meeting-style records and long-document QA.

\textbf{(iii) Multiple prompt variants support the family-level interpretation of BabelTele}. BabelTele does not rely on a single carefully optimized prompt. Instead, BabelTele-like prompts with different surface biases collectively trace a strong accuracy-retention frontier. This suggests that BabelTele is better understood as a family of model-readable high-density compressed representations rather than a single prompt trick.

\paragraph{Does Compression Make Models Think More?}

\begin{figure}[t]
    \centering
    \includegraphics[width=\columnwidth]{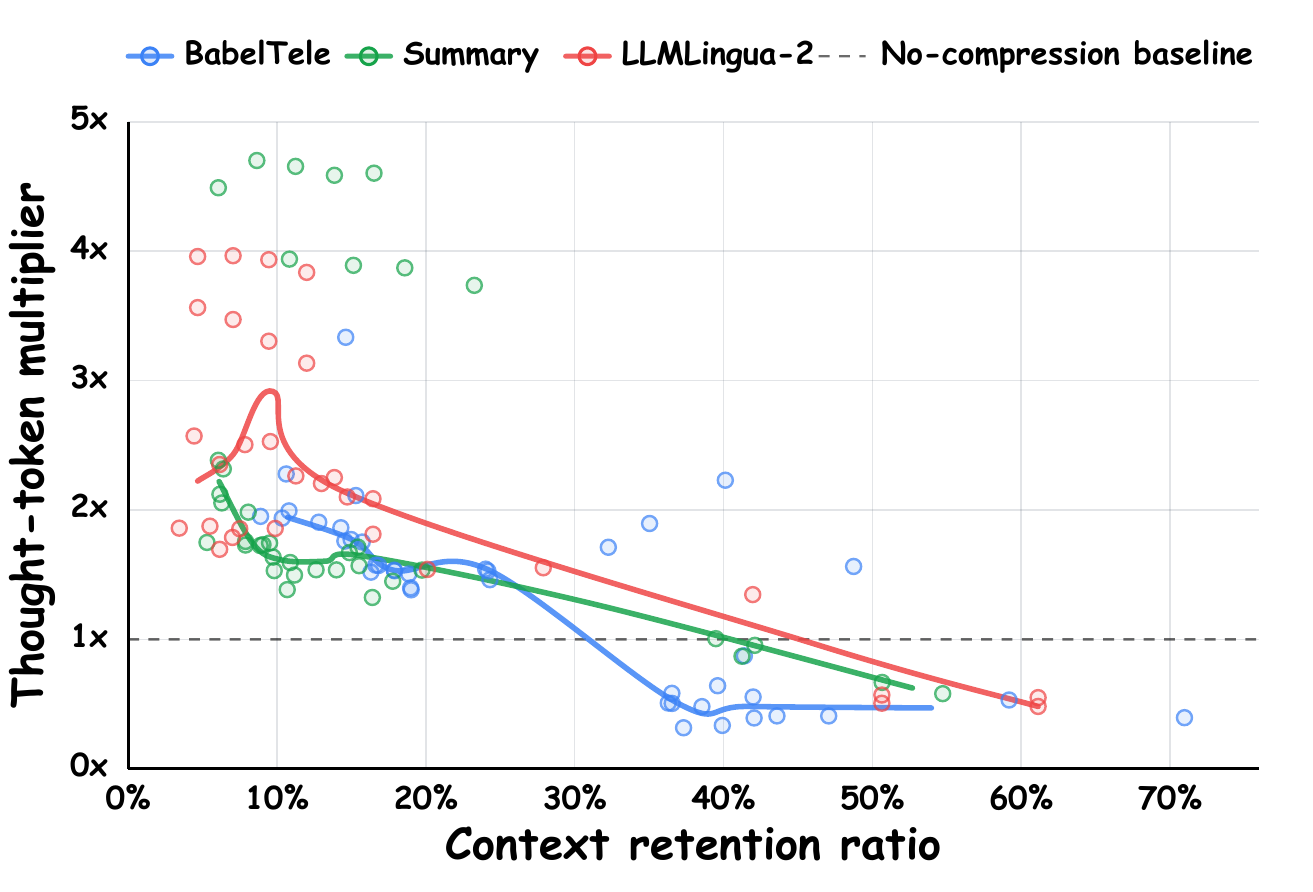}
    \caption{\textbf{Response chain-of-thought token multiplier versus realized context retention ratio}. Each marker corresponds to one compression run. Colors indicate compression methods, the y-axis is normalized by the corresponding no-compression baseline, and the horizontal dashed line marks $1\times$ chain-of-thought token usage. Solid colored curves show method-level smoothed trends.}
    \label{fig:chain-of-thought-token-retention}
    \vspace{-10pt}
\end{figure}

A natural concern is whether compression leads to longer response thought chains. In particular, one may ask whether the reader model needs to first decode BabelTele back into natural language before answering. Figure~\ref{fig:chain-of-thought-token-retention} examines the relation between context retention and response chain-of-thought token usage across all experimental settings, from which we summarize the following three points:

% \textbf{(i) Stronger compression often increases thought tokens.} As the context retention ratio decreases, the reader model tends to produce longer thought chains. This pattern appears across the evaluated compression methods, suggesting that increased chain-of-thought token usage is a general effect of compression rather than a phenomenon unique to BabelTele.

% \textbf{(ii) BabelTele does not necessarily introduce larger answer-side overhead.} Although BabelTele has a non-natural-language surface form, it induces a smaller overall chain-of-thought token increase than summary and LLMLingua-2 in many settings. In some relatively mild compression settings, its response chain-of-thought token multiplier is even lower than the original-context baseline, possibly because compression removes redundant or distracting context.

\textbf{(i) Stronger compression often increases chain-of-thought tokens}. As context retention decreases, chain-of-thought tokens generally rise across methods.

\textbf{(ii) BabelTele does not introduce unique overhead}. Its chain-of-thought token multiplier is often comparable to or lower than summary and LLMLingua-2, and can even fall below the original-context baseline in mild-compression settings.

\textbf{(iii) Thought-token growth reflects evidence accessibility}. A cautious interpretation is that longer thought chains arise when compression reduces the completeness or accessibility of retained evidence. If relevant details are removed or harder to use, the reader model may need more reasoning steps to reconstruct the answer or resolve uncertainty. Thus, BabelTele may introduce decoding cost, but because it often preserves task-relevant structure, its answer-side overhead is not necessarily higher than that of summary or LLMLingua-2.

\paragraph{Summary.}
Taken together, these results show that the model-side recoverability observed in Section 4.2 can translate into long-context compression gains. BabelTele exhibits a favorable accuracy-retention frontier across the QA settings, while the response chain-of-thought token analysis reveals an important efficiency boundary: extreme context compression triggers a space-time trade-off where input token savings are partially offset by CoT generation overhead. Importantly, this dynamic is intrinsic to LLM reasoning over sparse evidence rather than unique to BabelTele, suggesting that choosing an optimal, moderate compression ratio can yield simultaneous savings in both input context and reasoning steps.

%----------------------------------------------------------------------------------
\subsection{A Universal Cipher? Zero-Shot Cross-Model Comprehension}
To examine whether BabelTele represents a model-private shorthand or a transferable symbolic form, we evaluate its cross-model portability across state-of-the-art LLMs. We conduct compression-reading experiments on 180 samples from LongBench v2-Short and 214 QA instances from QuALITY, and analyze whether the effectiveness of BabelTele depends on specific compressor-reader pairs. For a more detailed discussion of experimental setup, please refer to Appendix~\ref{App_details_cross}.

%If BabelTele were merely a private shorthand of the model that produced it, its compressed texts should fail once they are read by a different model. The more interesting possibility is that BabelTele captures a shared, model-readable symbolic form: not fully universal, but portable enough that strong compressors can produce texts understood by heterogeneous LLMs. We therefore ask how far this portability extends, and whether it is uniform across models or structured by specific compressor-reader pairs.

%We conducted cross-model compression tests on several state-of-the-art large models using two QA subsets: 180 randomly selected samples from the Short subset of the LongBench v2 benchmark and 214 randomly selected question-answer instances from the QuALITY benchmark. The results can be analyzed from the following aspects:

\begin{figure}[t]
  \centering
  \includegraphics[width=\linewidth]{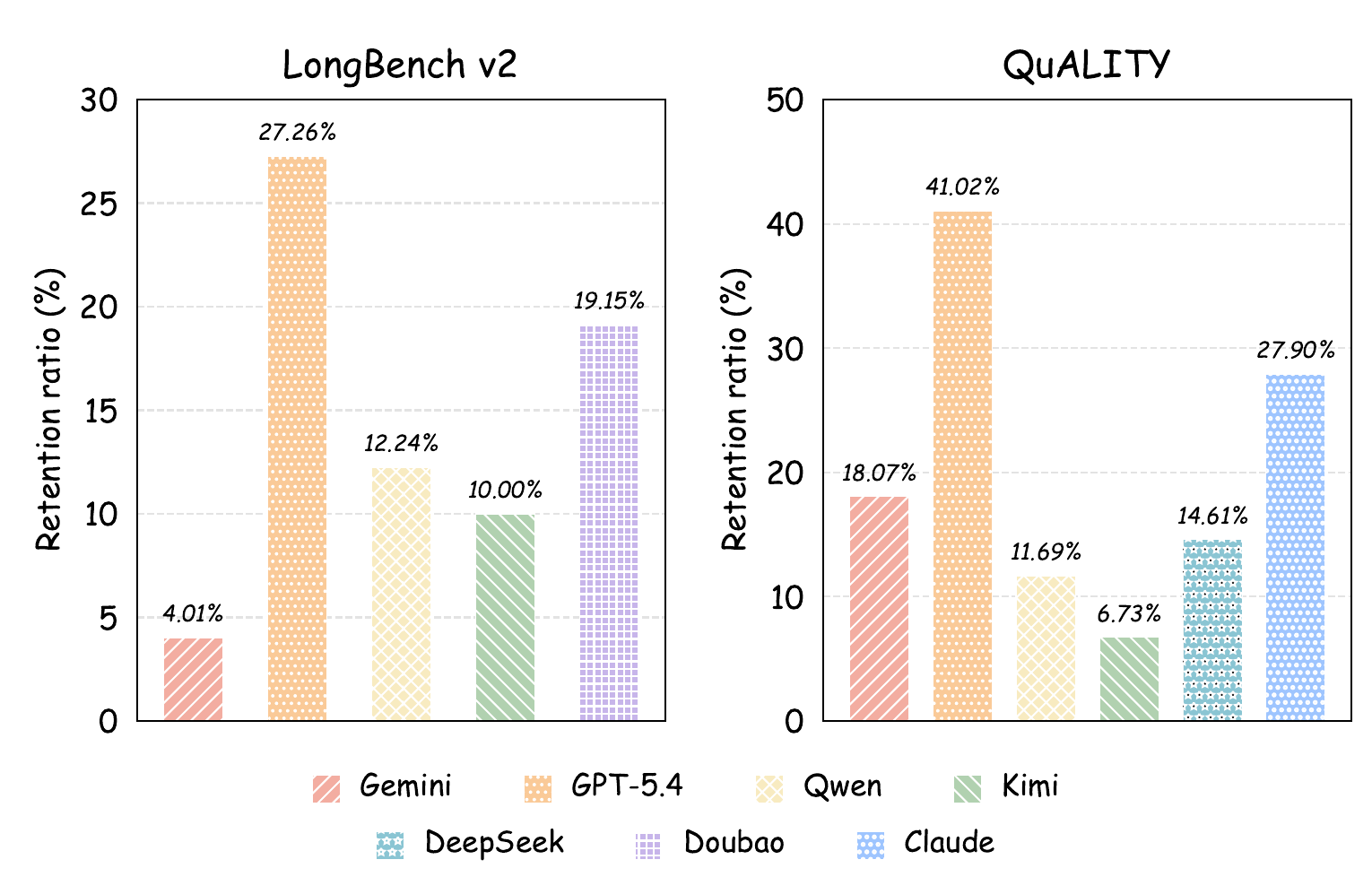}
  \vspace{-17pt}
  \caption{\textbf{Comparison of compression rates of different models}. We selected 180 samples from the Short subset of the LongBench v2 benchmark and processed them using BabelTele.}
  \label{Fig3}
  \vspace{-15pt}
\end{figure}
\textbf{(i) Compression Ratios of Different Models}. Figure~\ref{Fig3} shows that BabelTele compression strength varies substantially across compressors. Gemini 3.1 Pro is the most aggressive, exceeding \textbf{95}\% compression, whereas GPT-5.4 is more conservative at roughly \textbf{75}\%; the other models fall between \textbf{80}\% and \textbf{90}\%. This variation is important for interpreting the transfer matrices below, since higher compression can make the resulting symbolic form harder for other models to decode.

\begin{figure}[t]
  \centering
  \includegraphics[width=\columnwidth]{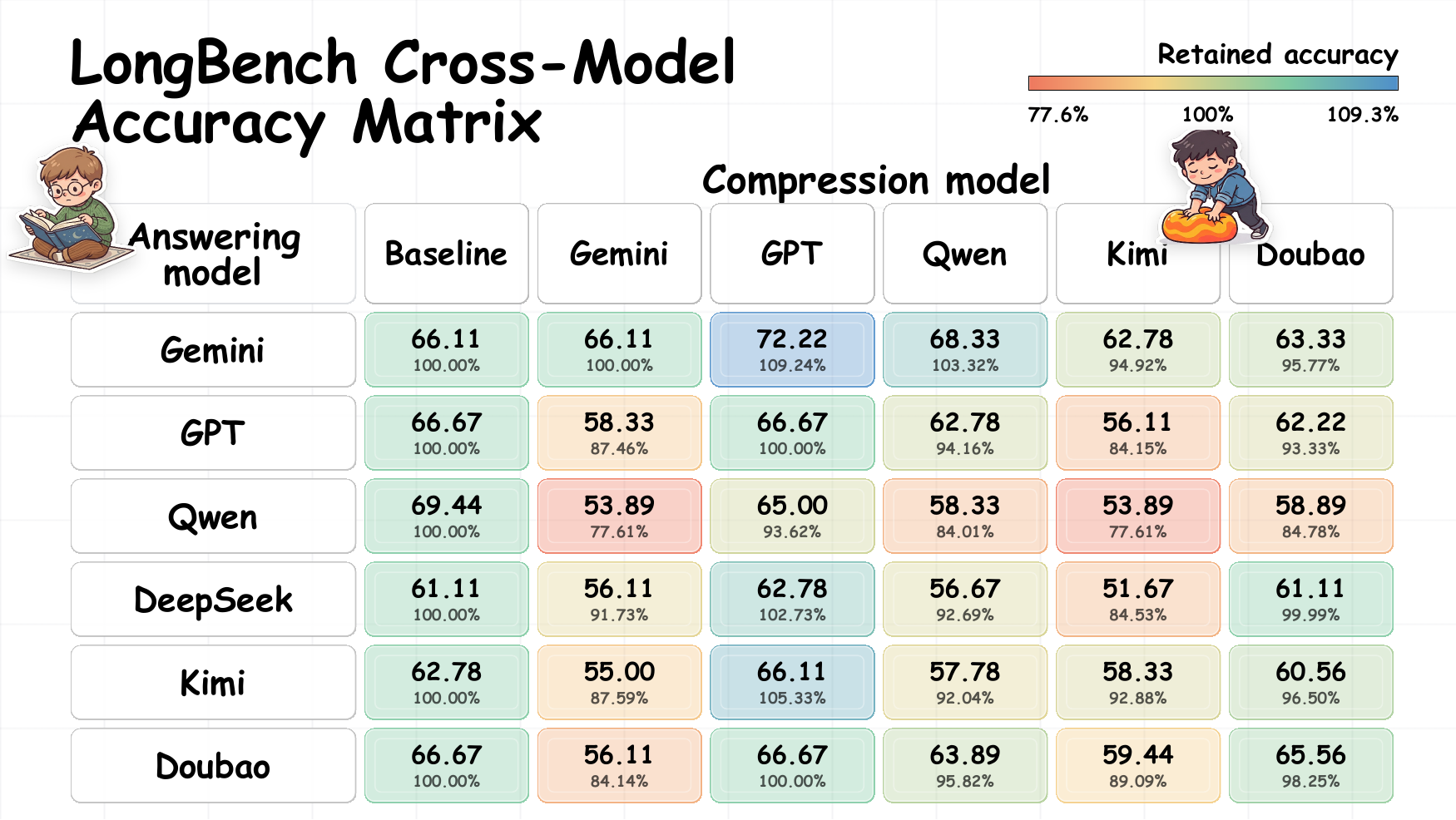}
  \caption{\textbf{Accuracy transfer matrix on 180 samples from the Short subset of LongBench v2}. Each row denotes answering models and columns denote compression models, with the Baseline column indicating no compression. Cell color summarizes retained performance relative to the no-compression baseline for the same answering model, and each cell reports absolute accuracy with retained performance shown below.}
  \label{fig:longbench-accuracy-transfer-matrix}
  \vspace{-10pt}
\end{figure}

\begin{figure}[t]
  \centering
  \includegraphics[width=\columnwidth]{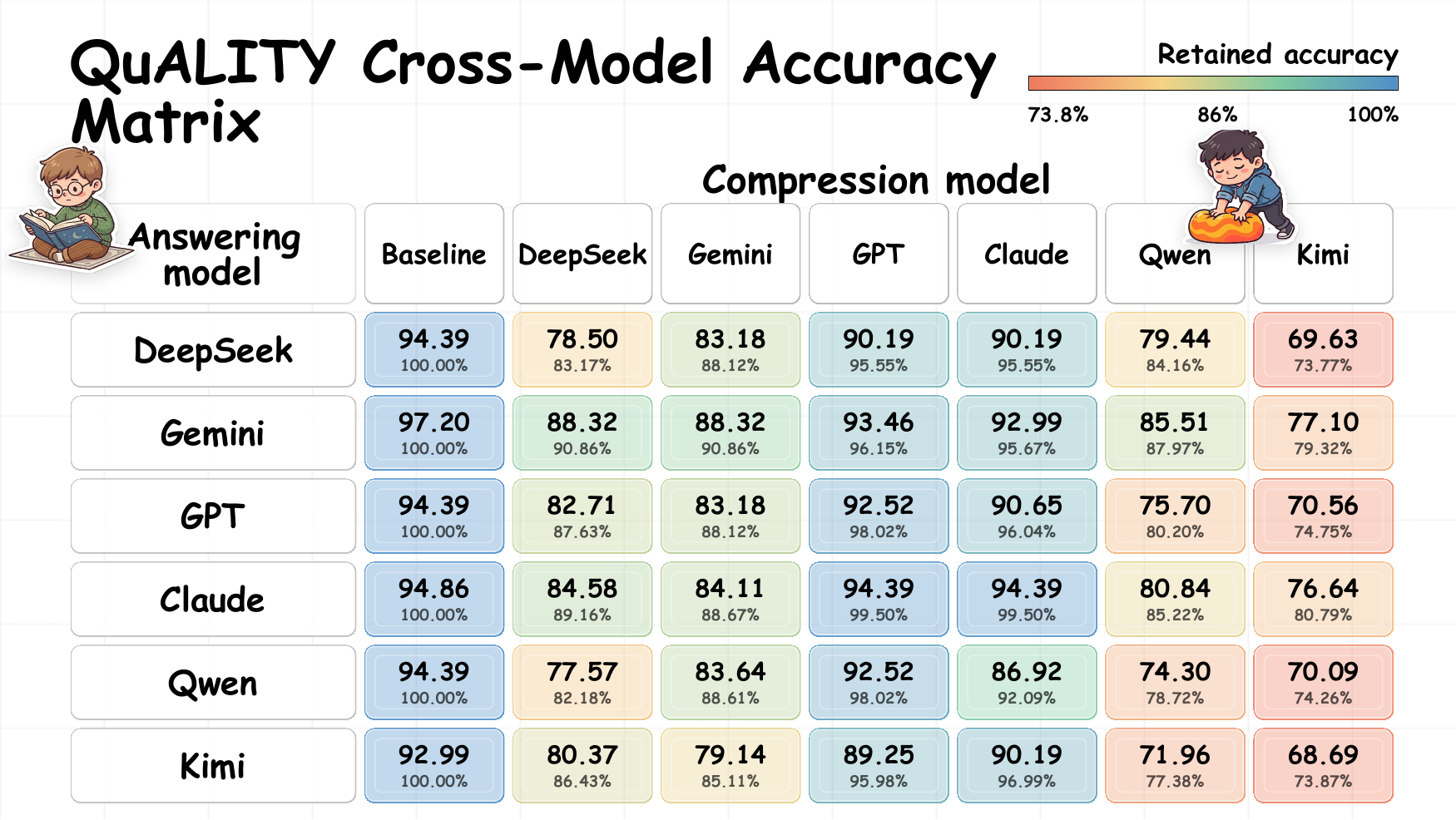}
  \caption{\textbf{Accuracy transfer matrix on QuALITY}. The matrix follows the same answering-model by compression-model layout and encoding as Figure~\ref{fig:longbench-accuracy-transfer-matrix}.}
  \label{fig:quality-accuracy-transfer-matrix}
  \vspace{-10pt}
\end{figure}

\textbf{(iI) Cross-Model Compression Accuracy}. Figures~\ref{fig:longbench-accuracy-transfer-matrix} and~\ref{fig:quality-accuracy-transfer-matrix} show that cross-model BabelTele comprehension is not dataset-specific. Across both LongBench v2 and QuALITY, compressed inputs remain usable for heterogeneous evaluators, but retention is strongly shaped by the compressor-reader pair. In particular, QuALITY shows that GPT-5.4 and Claude compressed inputs are broadly portable, while Qwen and Kimi compressed inputs lead to larger accuracy drops across readers. Thus, BabelTele is not a fully universal code, but its portability is systematic: strong compressors can produce symbolic forms that many models still understand.

%----------------------------------------------------------------------------------

\begin{figure}[t]
  \centering
  \includegraphics[width=\columnwidth]{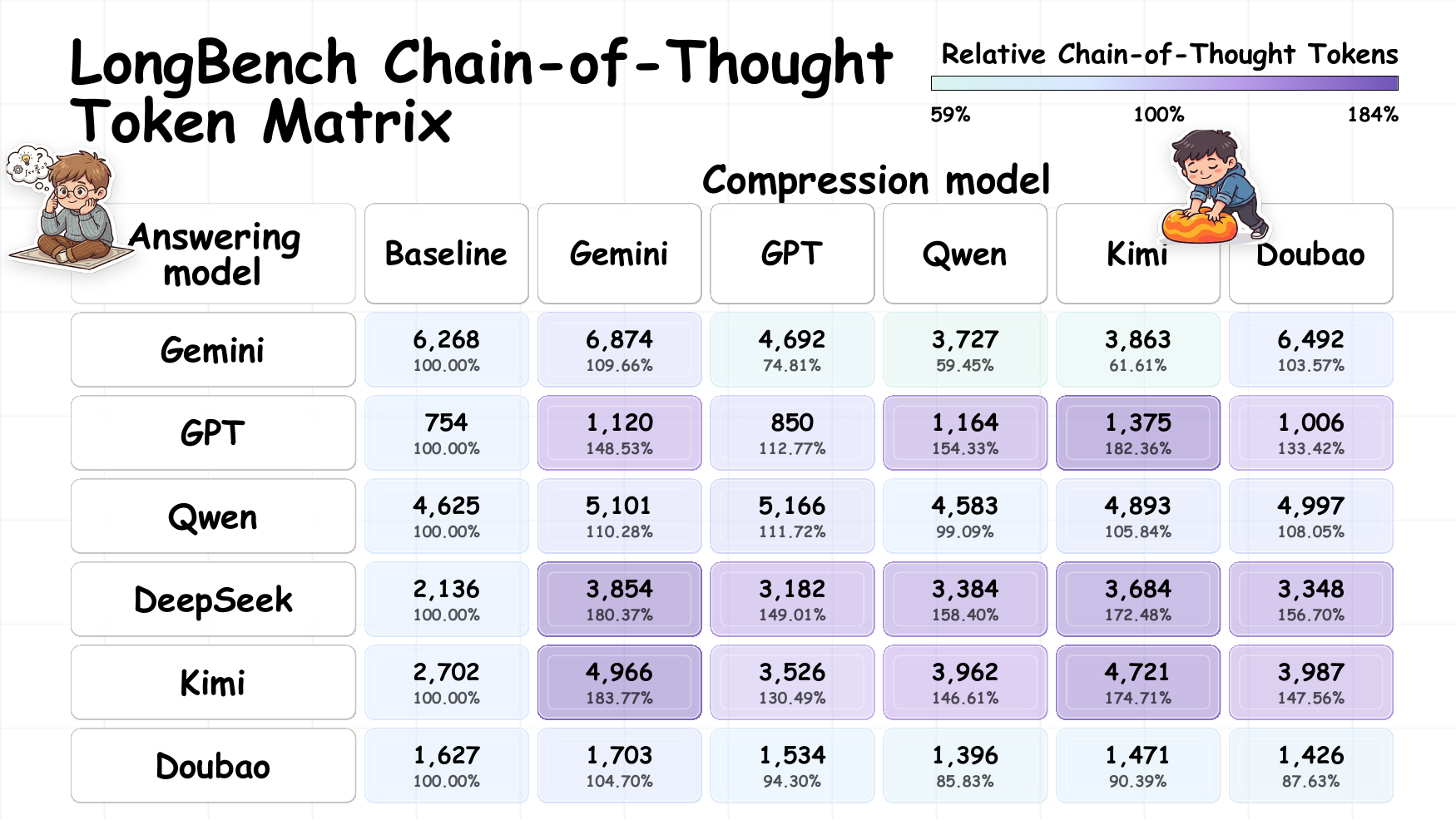}
  \caption{\textbf{Response chain-of-thought token transfer matrix on 180 samples from the Short subset of LongBench v2}. Rows denote answering models and columns denote compression models, with the Baseline column indicating no compression. Cell color summarizes the response chain-of-thought token ratio relative to the no-compression baseline for the same answering model, and each cell reports chain-of-thought tokens with relative ratio shown below.}
  \label{fig:longbench-chain-of-thought-token-transfer-matrix}
  \vspace{-10pt}
\end{figure}

\begin{figure}[t]
  \centering
  \includegraphics[width=\columnwidth]{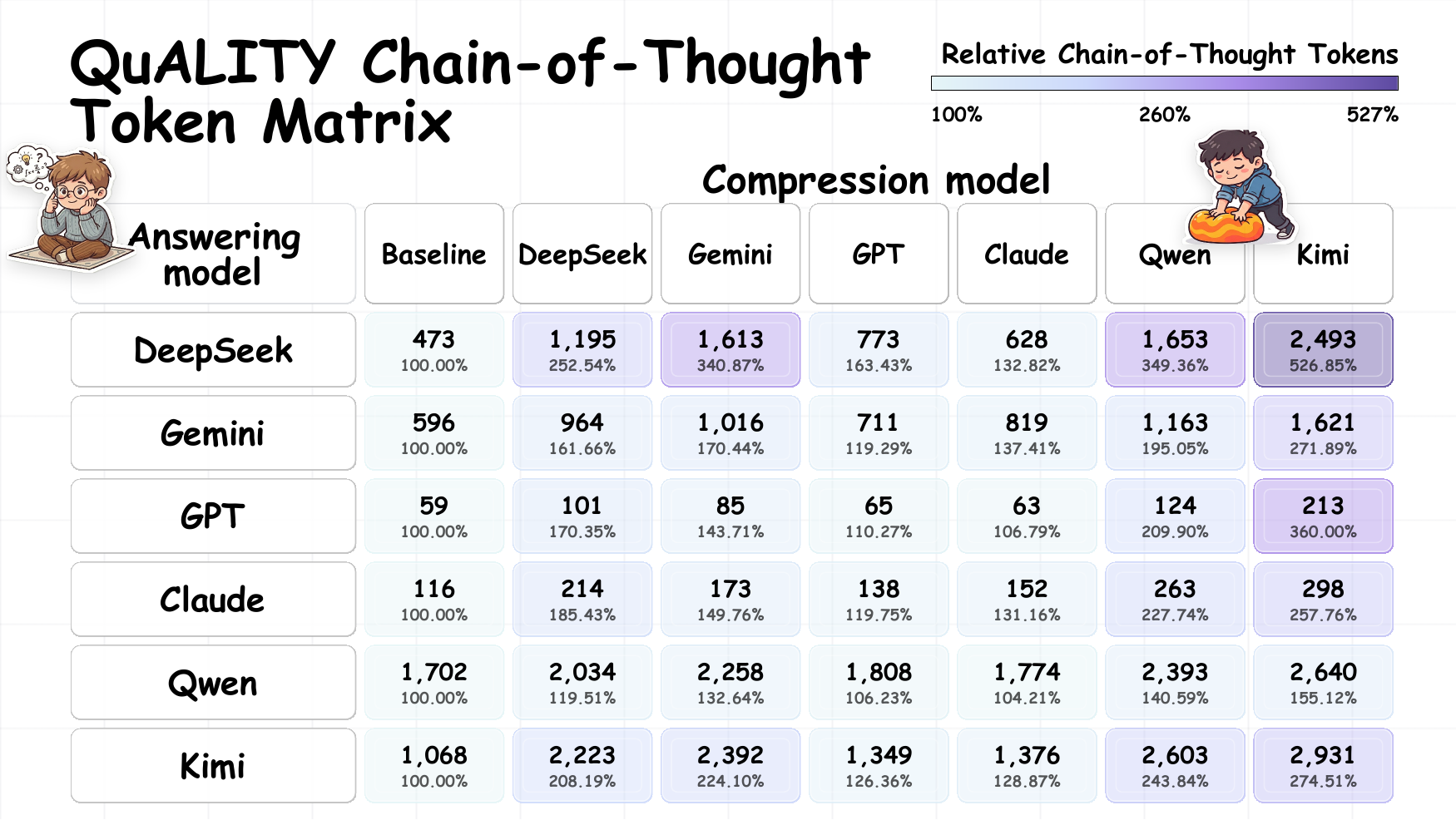}
  \caption{\textbf{Response chain-of-thought token transfer matrix on QuALITY}. The matrix follows the same answering-model by compression-model layout and visual encoding as Figure~\ref{fig:longbench-chain-of-thought-token-transfer-matrix}.}
  \label{fig:quality-chain-of-thought-token-transfer-matrix}
  \vspace{-6pt}
\end{figure}

\textbf{(iii) Cross-Model Inference Chain-of-Thought Length}. Figures~\ref{fig:longbench-chain-of-thought-token-transfer-matrix} and~\ref{fig:quality-chain-of-thought-token-transfer-matrix} report response thought tokens as a proxy for reader-side decoding overhead. Compressed inputs often increase this overhead, but the effect varies across evaluator-compressor pairs. This should be interpreted together with the compression-ratio results in Figure~\ref{Fig3}: more aggressive compressors may require reader models to spend more reasoning steps reconstructing or locating relevant evidence. Thus, longer thought chains do not necessarily indicate failed cross-model comprehension, but may partly reflect the general cost of higher compression. We therefore treat these values as auxiliary runtime evidence rather than direct evidence of semantic understanding.

\begin{table}[t]
\centering
\fontsize{8}{10}\selectfont
\setlength{\tabcolsep}{3pt}
\resizebox{\columnwidth}{!}{
\begin{tabular}{lcccc}
\toprule[1.5pt]
\textbf{Model} & \textbf{Origin} & \textbf{BabelTele} & \textbf{Drop} & \textbf{Retention} \\
\midrule
\rowcolor{green!3}
Qwen3-14B & 78.97\% & 64.02\% & -14.95 & 81.07\% \\
Qwen3.5-27B & 91.59\% & 80.84\% & -10.75 & 88.26\% \\
\rowcolor{green!3}
Qwen3.5-35B-A3B & 90.65\% & 75.70\% & -14.95 & 83.51\% \\
Qwen3.5-397B-A17B & 93.93\% & 79.91\% & -14.02 & 85.07\% \\
\rowcolor{green!3}
Qwen3.6-Max-Preview & 90.19\% & 79.44\% & -10.75 & 88.08\% \\
\bottomrule[1.5pt]
\end{tabular}
}
\caption{\textbf{Quality of Qwen-family models on original inputs and Gemini-induced BabelTele inputs}. The compressor is fixed to Gemini 3.1 Pro, while the evaluator varies across Qwen models. Drop is reported in percentage points.}
\label{tab:qwen-scale}
\vspace{-10pt}
\end{table}

\textbf{(iv) Scale Sensitivity within the Qwen Family}. To test whether BabelTele comprehension simply improves with model scale, we fix the compressor to Gemini 3.1 Pro and vary the Qwen-family evaluator. As shown in Table~\ref{tab:qwen-scale}, the Quality drop remains within a relatively narrow range, from \textbf{10.75} to \textbf{14.95} percentage points, and does not improve monotonically with model size. For example, Qwen3.5-397B-A17B has the highest original Quality but lower BabelTele Quality than Qwen3.5-27B. This suggests that BabelTele comprehension is not explained by scale alone, but also depends on model-specific robustness to the compressor-induced symbolic form.

\subsection{Capabilities and Boundaries in Downstream Tasks}

\subsubsection{Multi-Agent Communication}
We evaluate BabelTele representation under two multi-agent regimes: a homogeneous setting (both agents using Gemini 3.1 Pro) to  evaluate whether a model can produce and consume its own compression, and a heterogeneous setting (Gemini 3.1 Pro paired with GPT-5.4) to test cross-model portability as a black-box communication protocol.

Table~\ref{tab:multi-agent} presents the final results, from which we summarize the following two points: \textbf{(i) Significant token reduction}. BabelTele substantially reduces inter-agent communication overhead in both homogeneous and heterogeneous settings. This demonstrates that model-native compressed messages can effectively lower context consumption during repeated message passing, which is especially important for long-horizon multi-agent tasks. \textbf{(ii) Stable score maintaining}. Despite the strong compression, BabelTele maintains competitive final scores with only negligible performance degradation. This suggests that although the compressed messages are less readable to humans, they still preserve sufficient actionable information for LLM agents to coordinate and complete the task.

%原来的
%The main results are summarized in Table~\ref{tab:multi-agent}. In the homogeneous setting, BabelTele achieves significant compression of inter-agent communication while maintaining competitive performance. In the heterogeneous setting, BabelTele similarly reduces communication overhead with a compression ratio of \textbf{[X]}$\times$, with negligible impact on the final score. Although the compressed messages are opaque to human readers, they remain actionable for LLM agents and can be directly integrated into existing text-based agent frameworks without modifying model parameters or communication infrastructure. This suggests that model-native compression is a practical complement to existing multi-agent coordination methods, especially for long-horizon tasks where repeated message passing becomes a major source of context consumption.

\begin{table}[t]
    \fontsize{8}{13}\selectfont
    \centering
    \setlength{\tabcolsep}{12pt}
    \vspace{3pt}
    \begin{tabular}{lcc}
        \toprule[1.5pt]
        \textbf{Setting} & \textbf{Token Reduction} & \textbf{Score} \\
        \midrule
        \rowcolor{green!3}
        Homogeneous   & 38.96\% & 96.6\% \\
        \rowcolor{green!3}
        Heterogeneous & 44.21\% & 99.7\% \\
        \bottomrule[1.5pt]
    \end{tabular}
    \caption{\textbf{Performance of BabelTele in multi-agent communication settings}. Token Reduction denotes the proportion of tokens saved relative to uncompressed communication. Score is reported as a percentage of the uncompressed baseline.}
    \label{tab:multi-agent}
    \vspace{-8pt}
\end{table}

%----------------------------------------------------------------------------------

\subsubsection{Performance on Agent Memory}

We evaluate BabelTele on the representative LoCoMo agent memory benchmark using Gemini 3.1 Pro for compression and answering, with GPT-4o-mini as the evaluator. Full experimental details are provided in Appendix~\ref{App_details_locomem}.

%We evaluated the performance of BabelTele on the LoCoMo benchmark in the agent memory domain. This dataset consists of 10 conversations, each containing dozens of sessions. We independently compressed each session and generated a summary for each. Then, the query was compared with all summaries to compute similarity, and the top 4 most similar sessions were retrieved. All compression and answering were performed using the Gemini 3.1 Pro model, and the answers were evaluated with GPT-4o-mini model.

\begin{table}[t]
\fontsize{8}{13}\selectfont
\setlength{\tabcolsep}{7pt}
\centering
\vspace{3pt} 
\begin{tabular}{lcccc}
  \toprule[1.5pt]
  \textbf{Method}
       & \textbf{Token count ({\color{green!40!black}$\downarrow$})}
       & \textbf{Acc. ({\color{green!40!black}$\uparrow$})}
       & \textbf{\% ({\color{green!40!black}$\uparrow$})} \\
  \midrule
  \rowcolor{gray!15}
  Original
        &2819.5&64.81&100.00\% \\
  Summary
        &\textbf{1365.6}&61.05&94.20\% \\
  \rowcolor{green!3}
  BabelTele
        &1382.2&\textbf{62.53}&\textbf{96.48}\% \\
  \bottomrule[1.5pt]
\end{tabular}
\caption{\textbf{Performance on the LoCoMo benchmark}. \textbf{Token count} denotes the average total number of tokens consumed per query. The best result is highlighted in \textbf{bold black} font.}
\label{table4}
\vspace{-10pt} 
\end{table}

Table~\ref{table4} presents the final results, from which we can summarize the following three points: \textbf{(i) Robust memory retention}. Compared with the original uncompressed text, BabelTele representations incur minimal downstream accuracy loss, while preserving more actionable details than standard large language model summarization. \textbf{(ii) Lower compression rate}. Compared with other experiments, the compression rate of BabelTele on the LoCoMo benchmark is relatively low, around \textbf{50}\%. This is likely due to the short length of each session, which contains only about \textbf{700} tokens. Future work could explore experiments on larger agent memory benchmarks with longer sessions.

\subsubsection{Extending the Context Window}

We further evaluate BabelTele when the original input exceeds the model context window, using the Code Repo QA Long subset of LongBench v2. We compare direct truncation against BabelTele-based chunk compression across different models, with full details provided in Appendix~\ref{App_details_long_context}.

%We further evaluated model performance when the complete input text exceeds the model's context window. Specifically, we selected the Code Repo QA Long subset from LongBench v2 as the evaluation benchmark, whose average input length is approximately 1.65M tokens. The tested models include Qwen3.6 Max with a 256K context window, GLM-5.1 with a 200K context window, and Kimi2.6 with a 256K context window. For the original-input setting, we directly fed the complete text into the model and truncated the portion exceeding its context window. For BabelTele, we split the original text into chunks of 200K tokens, compressed each chunk using Gemini 3.1 Pro, concatenated the compressed outputs, and then fed the resulting text into the tested model.

\begin{table}[t]
\fontsize{8}{13}\selectfont
\setlength{\tabcolsep}{7pt}
\centering
\vspace{3pt} 
\begin{tabular}{lcccc}
  \toprule[1.5pt]
  \textbf{Method}
       & \textbf{Qwen3.6 Max}
       & \textbf{GLM-5.1}
       & \textbf{Kimi2.5} \\
  \midrule
  \rowcolor{gray!15}
  Original
        &55.17&62.07&44.82 \\
  \rowcolor{green!3}
  BabelTele
        &\textbf{62.07}&\textbf{72.41}&\textbf{48.28} \\
  \bottomrule[1.5pt]
\end{tabular}
\caption{\textbf{Performance on the Code Repo QA Long subset from LongBench v2 benchmark}. The best result is highlighted in \textbf{bold black} font.}
\label{table5}
\vspace{-10pt} 
\end{table}

Table~\ref{table5} presents the results. On Qwen3.6-Max, the original truncated input achieves an accuracy of \textbf{55.17}\%, while the  dense BabelTele representations allow the model to capture broader evidence, achieving an accuracy of \textbf{62.07\%}. This result indicates that when the original text exceeds the model's context window, direct truncation discards a large amount of potentially important information, thereby impairing the model's understanding and reasoning ability. In contrast, BabelTele effectively condenses the core information from ultra-long texts, allowing the model to receive more complete content within its limited context window and thus mitigating the limitations caused by insufficient context capacity.

\section{Conclusion}

This paper investigates BabelTele, a high-density textual representation optimized for model decodability rather than human readability. Our experiments demonstrate that BabelTele achieves strong compression ratios with negligible performance degradation, while remaining semantically recoverable by LLMs. Notably, this capability generalizes across a diverse set of proprietary and open-weight models in a zero-shot manner, suggesting that the ability to interpret such representations is a general capability of LLMs rather than an artifact of any particular model. In practical scenarios including multi-agent communication and agent memory, BabelTele shows promising potential as a model-native intermediate representation. We therefore view BabelTele not as a finished protocol, but as evidence that high-density textual representations optimized for LLM-to-LLM communication need not prioritize human readability, and as a direction worth further exploration.

\section*{Limitations}

Our current evaluation focuses on a selected set of benchmarks and model families, the behavior of BabelTele across a broader range of tasks and architectures remains to be explored. Additionally, as an empirical study, this work primarily characterizes the phenomenon rather than explaining its underlying mechanisms, a deeper theoretical understanding of how LLMs form and interpret model-native representations is left to future work.

% Our methodology relies on zero-shot prompting, meaning the generated representations are stochastic and may vary across runs or models, making BabelTele difficult to deploy as a consistent protocol in practice. As our study is purely empirical, it remains unclear under what conditions BabelTele reliably preserves meaning. Future work may explore fine-tuning or reinforcement learning to stabilize the generation process, as well as information-theoretic analysis to better characterize the semantic boundaries of LLM-native representations.

\section*{Ethics Statement}
We acknowledge that all authors are informed about and adhere to the ACL ARR Code of Ethics and the Code of Conduct.

\subsection*{Risks}
Our benchmarks are sourced from publicly available datasets. We cannot guarantee that they are free of biased, toxic, or otherwise harmful content. In addition, BabelTele transforms text into a compact, non-standard representation, which may alter the behavior of the original text in unexpected ways; when applied to safety-critical domains, such changes could compromise safety or introduce unintended risks. We used LLM-based AI tools only for grammar and language polishing; all technical content, experiments, and claims were written and verified by the authors.

% This document does not cover the content requirements for ACL or any
% other specific venue.  Check the author instructions for
% information on
% maximum page lengths, the required ``Limitations'' section,
% and so on.

% This document has been adapted
% by Steven Bethard, Ryan Cotterell and Rui Yan
% from the instructions for earlier ACL and NAACL proceedings, including those for
% ACL 2019 by Douwe Kiela and Ivan Vuli\'{c},
% NAACL 2019 by Stephanie Lukin and Alla Roskovskaya,
% ACL 2018 by Shay Cohen, Kevin Gimpel, and Wei Lu,
% NAACL 2018 by Margaret Mitchell and Stephanie Lukin,
% Bib\TeX{} suggestions for (NA)ACL 2017/2018 from Jason Eisner,
% ACL 2017 by Dan Gildea and Min-Yen Kan,
% NAACL 2017 by Margaret Mitchell,
% ACL 2012 by Maggie Li and Michael White,
% ACL 2010 by Jing-Shin Chang and Philipp Koehn,
% ACL 2008 by Johanna D. Moore, Simone Teufel, James Allan, and Sadaoki Furui,
% ACL 2005 by Hwee Tou Ng and Kemal Oflazer,
% ACL 2002 by Eugene Charniak and Dekang Lin,
% and earlier ACL and EACL formats written by several people, including
% John Chen, Henry S. Thompson and Donald Walker.
% Additional elements were taken from the formatting instructions of the \emph{International Joint Conference on Artificial Intelligence} and the \emph{Conference on Computer Vision and Pattern Recognition}.

% \bibliography{custom}
% Re-enable the bibliography once the draft contains active citations.

% \bibliographystyle{acl_natbib}
\bibliography{custom}

\clearpage
\appendix
\addtocontents{toc}{\protect\appendixTOCstart}
\renewcommand{\contentsname}{Appendix}

\begingroup
\setcounter{tocdepth}{3}
\makeatletter

\let\oldl@subsection\l@subsection
\renewcommand{\l@subsection}[2]{%
  \vspace{0.8ex}
  \oldl@subsection{#1}{#2}%
}

\let\oldl@subsubsection\l@subsubsection
\renewcommand{\l@subsubsection}[2]{%
  \vspace{0.8ex}
  \oldl@subsubsection{#1}{#2}%
}

\makeatother
\tableofcontents
\endgroup

\setcounter{paragraph}{0}
\renewcommand{\theparagraph}{}

\section{Related Works}
\label{App_relatedworks}

\subsection{Learned and Latent Context Compression}
Another line of work compresses context into learned tokens, memory vectors, or internal activations \citep{mu2023gist, chevalier2023autocomp, ge2024icae, zhang2025actbeacon, liang2025ilre, hooper2024kvquant, li2026latent}. Gist Tokens \citep{mu2023gist} summarize prompts into reusable special tokens, while AutoCompressors \citep{chevalier2023autocomp} compress long contexts into compact summary vectors as soft prompts.
BabelTele differs in its interface assumption: unlike learned-token or activation-level methods that often require training, hidden-state access, special tokens, or architectural changes, it produces discrete text usable through black-box LLM APIs, while not being constrained to natural language.

\subsection{Memory, Retrieval and Long-Context LLM Systems}
Long-context LLM applications also motivate compact memory and retrieval representations \citep{lewis2020rag, yoran2024ragrobo, bai2024longbench, sung2023genagent, packer2023memgpt, xu2025amem, yu2025memagent, borro2026memori, maharana2024locomo}. Retrieval-augmented generation prepends external documents to the prompt, but retrieved passages are often verbose and noisy \citep{lewis2020rag, yoran2024ragrobo, xu2024recomp}. Contextual compression for RAG reduces this burden by filtering, summarizing, or restructuring retrieved evidence \citep{xu2024recomp}.
LLM agents introduce a related memory bottleneck. Generative Agents \citep{packer2023memgpt} maintain a natural-language memory stream with reflection and retrieval. MemGPT \citep{yu2025memagent} manages working context and external memory through an OS-like memory hierarchy.
BabelTele is complementary to these systems: rather than changing the retrieval or memory controller, it proposes a denser representation format for information that will mainly be consumed by LLMs.

\subsection{Symbolic Representations and LLM-Native Communication}
BabelTele is related to symbolic and machine-to-machine communication \citep{foerster2016deepmarl, havrylov2017emerlan, mordatch2018emerground, yin2023exchange, marro2024scalable, ramesh2025commact, zou2025latentcol, du2025enablelatent, gassen2026sematiccomp, gao2023pal, jiang2023structgpt, schnabel2024symbprompt}. Emergent communication studies non-human-readable protocols, while Exchange-of-Thought \citep{yin2023exchange} and Agora \citep{marro2024scalable} explore reasoning-trace exchange and scalable communication among LLM agents.
Symbolic prompting is another nearby direction. 
MetaGlyph \citep{gassen2026sematiccomp} compresses instructions with symbolic metalanguages, while structured prompting uses non-natural-language formats such as code, JSON, and tables \citep{gao2023pal, jiang2023structgpt, schnabel2024symbprompt}.
BabelTele differs in that it does not rely on a manually designed symbolic language or a fixed schema. Instead, it studies whether LLMs can be prompted to invent compact, LLM-readable encodings for arbitrary semantic content.

\section{Experimental Setup}
\label{App_setup}

\subsection{Task-Agnostic Compression Protocol}
For document QA experiments, BabelTele compression is performed in advance before downstream questions are introduced. The compressor receives only the source passage or document context, and does not observe questions, answer options, gold answers, or evaluation prompts.

\subsection{Baselines}
We compare BabelTele against the original uncompressed context, natural-language summaries, and LLMLingua-2~\citep{pan2024llmlingua2} under matched settings. These conditions separate no-compression performance, human-readable summarization, and learned prompt compression from BabelTele-style model-oriented compression.

\subsection{Datasets}
We evaluate BabelTele on both intrinsic compression diagnostics and downstream task performance.

\paragraph{QuALITY.}
QuALITY \citep{pang2022quality} is used as a long-document multiple-choice QA benchmark. In the pilot setting, we sample 10 long passages, each paired with 3 questions, producing 30 question-answer instances. For each passage, we construct three context variants: the original passage, a natural-language summary, and a BabelTele-compressed representation. In the larger QuALITY evaluation, we further compare BabelTele with LLMLingua-2 and report results by source domain, passage length, and question hardness.

\paragraph{LongBench v2.}
LongBench v2 \citep{bai2025longbenchv2} is used to test long-context document QA under stronger scale and cross-model transfer conditions. We evaluate a subset of 180 samples under no compression and multiple BabelTele compression sources. This setting allows us to separate two factors: the model that produces the compressed representation and the model that reads it.

\paragraph{LoCoMo.}
LoCoMo \citep{maharana2024locomo} is used as an initial testbed for long-term conversational memory compression. Instead of compressing isolated passages, this setting compresses dialogue histories or memory contexts before answering memory dependent questions. The goal is to test whether BabelTele can reduce agent memory storage while preserving enough reliable information for downstream recall.

\paragraph{DeepResearch.}
We built a multi-agent system consisting of two agents and tested it on the DeepResearch Bench \citep{du2025deepresearch}, to evaluate BabelTele in multi-agent communication. In this setting, intermediate messages between agents are compressed before being passed to the next agent.

\paragraph{MeetingBank.}
MeetingBank \citep{hu2023meetingbank} is used to test long-context summarization and question answering on real-world spoken dialogue. We sample a subset of city council meeting transcripts to evaluate the models' ability to process verbose, multi-party conversations. In this setting, the extensive meeting records are compressed before being passed to the target model for downstream tasks. Because MeetingBank serves as the primary training corpus for LLMLingua-2, this evaluation allows us to directly benchmark BabelTele against LLMLingua-2 in the baseline's native domain.

\subsection{Models}
Our experiments cover both compression models and reader models. For pilot generation, we use Gemini 3.1 Pro \citep{google2026gemini3-1} as the compressor. For cross-model evaluation, we include models from mainstream proprietary and open-weight families, including GPT-5.4 \citep{openai2026gpt5-4}, Kimi K2.5 and Kimi K2.6 \citep{kimi2026kimi2-5, kimi2026kimi2-6}, Meta Llama 3 8B \citep{llama2024llama3}, Qwen2-7B \citep{qwen2}, Qwen2.5-7B \citep{qwen2.5}, Qwen3-8B, Qwen3-14B, Qwen3-32B \citep{qwen3}, DeepSeek-V4-Pro \citep{deepseekai2026deepseekv4}, GLM-5.1 \citep{glm5team2026glm5}, Qwen3.6-Plus \citep{qwen36plus}, DeepSeek-R1 \citep{deepseek2025r1}, Doubao-Seed-2.0 \citep{seed2}, Claude Sonnet 4.6 \citep{claude2026sonnet46}, Qwen3.5-27B, Qwen3.5-35B-A3B, Qwen3.5-Plus, Qwen3.5-397B-A17B \citep{qwen3.5}, Qwen3.6-Max-Preview \citep{qwen36_max_preview}.

\subsection{Metrics.}
We evaluate BabelTele along four dimensions.

\paragraph{Compression.}
We report token count and context retention ratio as basic compression statistics. The retention ratio is the length of the compressed context divided by the length of the original context, so lower values indicate stronger compression.

\paragraph{Semantic Fidelity.}
We use downstream QA accuracy as the primary semantic fidelity metric to assess answer preservation. For each compressed context, the reader model answers the same questions as in the original-context setting. We also report normalized accuracy, defined relative to the no-compression condition, to show how much task performance is retained after compression.

\paragraph{Human Readability.}
To characterize the surface form of BabelTele from complementary perspectives, we compute readability, out-of-vocabulary ratio, cross-entropy, and perplexity under several language models. We conduct a human questionnaire on a subset, measuring human QA accuracy, perceived difficulty, and completion time.

\paragraph{System Utility.}
For agent memory and multi-agent communication, we focus on operational metrics in realistic interactive settings: context token reduction, task accuracy or success rate, and, where available, response or reasoning-token overhead. These metrics connect intrinsic compression to practical system-level benefits.

\section{Prompt Templates}
\label{App_prompts}

\subsection{BabelTele Compression Prompt}
\label{App_prompt_default}

The following prompt is used to elicit BabelTele representations from the compressor model. The source passage or document context is appended after the final line of the prompt.
Unless otherwise specified, this is the default compression prompt used in most experiments in this paper, except for the Section~\ref{section4_3} prompt-family sweep.

\begin{PromptBox}
your task: compress verbose human text into minimal Token sequence. Audience \ensuremath{\neq} human, but another equally intelligent LLM.
Core Directive
Omnilingual: ignore single-language grammar; traverse all human languages (Chinese, English, German compounds, Japanese Kanji, Latin roots, etc.), pick highest info-density words.
Symbolic Collapse: optionally replace conjunctions, emotions, long sentences with Emoji, math/logical symbols (=>, \ensuremath{\in}, \ensuremath{\neq}), punctuation.
Universality: any LLM should fully understand compressed output without a codebook.
Lossless: retain all information & details.

Compress the content bellow:
\end{PromptBox}

\input{babeltele_prompt_family_appendix}

\section{Qualitative Document-Level Example}
\label{App_examples}

\begin{figure*}[p]
\centering
\includegraphics[width=\textwidth]{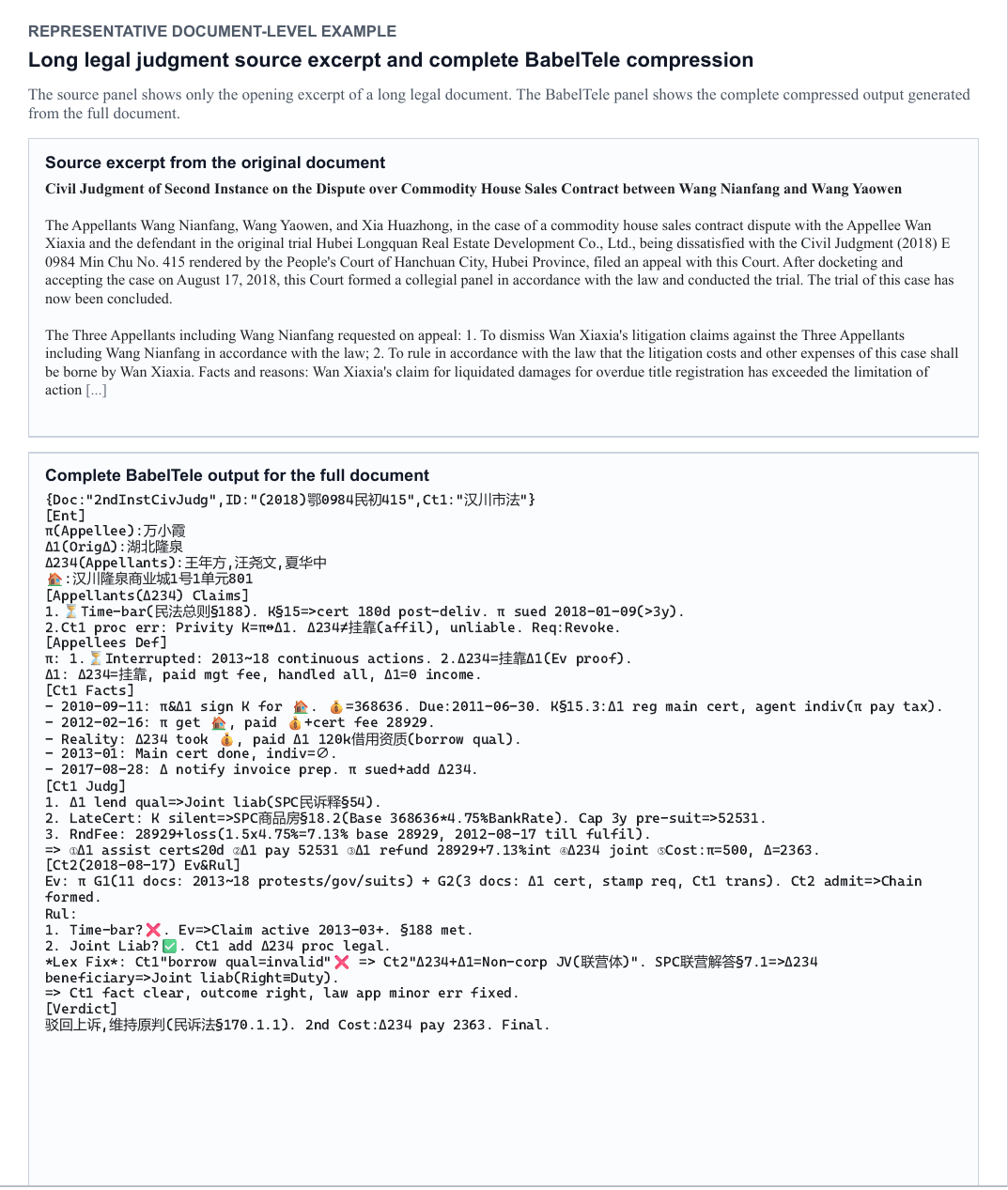}
\caption{\textbf{Representative document-level BabelTele example}. The source excerpt indicates the genre and information density of the original legal document; the BabelTele panel shows the complete compressed output generated from the full document.}
\label{fig:babeltele-document-example}
\end{figure*}

Because BabelTele is primarily applied to long documents, displaying complete source documents is impractical in the paper format. Figure~\ref{fig:babeltele-document-example} provides a representative document-level compression example. The source panel shows only the opening excerpt of a long legal judgment, while the BabelTele panel shows the complete compressed representation generated from the full document. This example is intended to illustrate the surface form of BabelTele rather than serve as additional quantitative evidence.

\clearpage

\section{Implementation Details}

\subsection{Symbolic Collapse: Separating Human Readability from Model Decodability}
\label{App_details_symbolic}

In early exploratory observations and small-scale informal tests of BabelTele, the most interesting and intuitive phenomenon was that although this representation is often difficult for humans to read directly, large language models (LLMs) still seem capable of using it to answer questions, recover details, and even perform a certain degree of reasoning. Inspired by this phenomenon, we first examine whether human readability, natural-language distribution typicality, and the model's ability to recover semantics can be experimentally decoupled. This section does not focus on the efficiency of BabelTele as a compression method, but rather on whether it pushes text outside the realm of human-readable natural language while still preserving semantic structures usable by LLMs.

We select 10 long-text samples from the QuALITY dataset, with each sample containing 3 multiple-choice question-answering (QA) items. For each sample, we construct three input formats: the original text, a human-oriented natural-language summary, and the BabelTele representation. The original text represents standard natural-language input; the natural-language summary serves as a readable paraphrased reference to control for the factor of ``text being rewritten or shortened by models''; and BabelTele represents a model-oriented representation that deliberately abandons human readability. Both the summary and BabelTele representations were generated by Gemini 3.1 Pro, and all three formats were evaluated on the same set of questions. To mitigate the variance introduced by the stochastic nature of LLM generation, all experiments in Section 4.2 were repeated three times, and the reported results are the averages across these independent runs.

To test this separation, we combine surface readability diagnostics, distributional likelihood metrics, and behavioral QA evaluations from both humans and LLMs. These measurements evaluate semantic recoverability in downstream tasks, rather than making a direct claim that the model ``understands'' BabelTele in a human-like sense.

First, we evaluate the surface readability of the three text formats using the Dale-Chall Readability Score and the corresponding proportion of difficult words. Dale-Chall is sensitive to uncommon lexical forms, abbreviations, proper nouns, and code-like tokens, which are central to the surface form of BabelTele. Since BabelTele contains multilingual elements, symbols, abbreviations, and non-standard structures, this metric should be interpreted as a surface-form diagnostic rather than a complete measure of human comprehension. Nevertheless, it provides an automated reference for whether BabelTele deviates from conventional natural-language text. As shown in Table~\ref{tab:symbolic-collapse-readability}, BabelTele obtains higher Dale-Chall scores and difficult-word ratios than both the original and summary texts.

\begin{table}[t]
\fontsize{8}{13}\selectfont
\setlength{\tabcolsep}{7pt}
\centering
\vspace{3pt}
\begin{tabular}{lcccc}
  \toprule[1.5pt]
  \textbf{Variant} & \textbf{$n$} & \textbf{Dale-Chall} & \textbf{Difficult words} \\
  \midrule
  \rowcolor{gray!15}
  Original & 10 & 10.28 & 35.97\% \\
  Summary & 10 & 13.51 & 56.34\% \\
  \rowcolor{green!3}
  BabelTele & 10 & \textbf{16.70} & \textbf{80.19}\% \\
  \bottomrule[1.5pt]
\end{tabular}
\caption{\textbf{Dale-Chall readability diagnostics for the Original / Summary / BabelTele triplets}. Higher scores and larger difficult-word ratios indicate lower surface readability under this English-prose readability metric. The best result is highlighted in \textbf{bold black font}.}
\label{tab:symbolic-collapse-readability}
\vspace{-10pt}
\end{table}

Second, we input the original, summary, and BabelTele texts into language models to calculate PPL, BPB, and BPC. Here, PPL (Perplexity) measures how difficult the text is for a language model to predict; BPB (Bits Per Byte) measures the average amount of information required per byte; and BPC (Bits Per Character) measures the average information complexity per character. These metrics do not directly measure whether the text is understandable; instead, they reflect the prediction difficulty of the text sequences under the language model's distribution. Therefore, they are used to determine whether BabelTele is a low-likelihood surface form that lies outside the general distribution of natural language. As shown in Table~\ref{tab:symbolic-collapse-distribution}, BabelTele yields substantially higher PPL than the original and summary texts across multiple base language models.

Finally, we compare the performance of human readers and LLMs on the QA tasks. For the LLM evaluation, we feed the different input formats, along with their corresponding multiple-choice questions, into Gemini 3.1 Pro and prompt the model to return only the option indices. For the human evaluation, we construct questionnaires asking subjects to read either the original or BabelTele text and answer the corresponding questions, while recording their subjective difficulty ratings and QA performance. Since the human questionnaires primarily compare the original and BabelTele conditions, Figure~\ref{fig:symbolic-collapse-qa-accuracy} focuses on the same two conditions for both humans and Gemini 3.1 Pro.

First, the results indicate that BabelTele is no longer natural language in the conventional sense for humans. Dale-Chall diagnostics in Table~\ref{tab:symbolic-collapse-readability} show that BabelTele scores higher than both the original and summary texts in terms of readability score and difficult-word ratio. This indicates that it contains a large number of abbreviations, symbols, proper nouns, and non-standard expressions that English readability models struggle to handle. Human questionnaires also reveal that subjects in the BabelTele condition exhibit lower QA accuracy and report higher subjective difficulty, demonstrating that this low readability is reflected not only in automated metrics but also in practical semantic recovery tasks, as shown in Figure~\ref{fig:symbolic-collapse-qa-accuracy}.

For the models, BabelTele likewise resembles an out-of-distribution representation for natural language. Taking Llama-3-8B as an example, Table~\ref{tab:symbolic-collapse-distribution} shows that the average PPL of the original and summary texts is approximately 9.63 and 11.32, respectively, whereas the PPL for BabelTele surges to around 176.60. Similar trends are observed across multiple Qwen base models. This demonstrates that BabelTele is not a simple summary or shorthand of ordinary natural language; rather, its surface form significantly deviates from the natural-language distribution. It should be emphasized that high PPL measures the low likelihood of the sequence, not semantic unrecoverability.

However, the aforementioned two types of deviations do not result in the failure of the model's semantic recoverability. On the same QuALITY QA task, Figure~\ref{fig:symbolic-collapse-qa-accuracy} shows that Gemini 3.1 Pro maintains high accuracy when using BabelTele, without exhibiting the significant performance collapse observed in the human questionnaires. This demonstrates that BabelTele differs from meaningless gibberish: although it reduces human readability and is highly atypical under base model distributions, it still retains sufficient entity, relational, event, and detailed information for use by instruction-tuned LLMs. This indicates that human readability, natural-language distribution typicality, and model semantic recoverability can be decoupled.

Given that BabelTele is not meaningless gibberish, but a representation characterized by low human readability and low natural-language likelihood while remaining decodable by models, can this semantic recoverability be translated into compression gains in real-world long-context tasks?

\subsection{Efficiency and Cognitive Overhead in Model-Native Compression}
\label{App_details_native}

\paragraph{Dataset and Evaluation Format.}
QuALITY tests fine-grained evidence preservation in long-document reading comprehension. MeetingBank is converted into a multiple-choice QA format, respectively, so that both benchmarks can be evaluated using a consistent unified accuracy metric.

\paragraph{Retention Sweep Construction.}
Since generative compression methods cannot precisely control their realized compression ratios, we do not compare methods at a single nominal compression point. Instead, we evaluate fairer accuracy-retention curves. LLMLingua-2 forms a sweep by specifying different target compression ratios, while the summary baseline forms an empirical sweep by prompting the model to approximately compress to different target ratios.

\paragraph{BabelTele Prompt Variants.}
For BabelTele, we use multiple BabelTele-like prompts that all instruct the model to abandon ordinary human readability while preserving task-relevant semantics. These variants introduce different surface biases, including multilingual mixing, logical symbols, structured tags, and entity-relation compression. They naturally produce different realized retention ratios, forming a BabelTele retention sweep and allowing us to test whether the observed effect reflects a broader family of model-readable high-density representations rather than a single prompt artifact. Full prompt texts are provided in Appendix~\ref{App_prompt_family_native}.

\subsection{A Universal Cipher? Zero-Shot Cross-Model Comprehension}
\label{App_details_cross}

If BabelTele were merely a private shorthand of the model that produced it, its compressed texts should fail once they are read by a different model. A more interesting possibility is that BabelTele captures a shared, model-readable symbolic form: although it is not fully universal, it may be sufficiently portable for strong compressors to produce compressed texts that can be understood by heterogeneous LLMs. Therefore, we investigate how far this portability extends and whether it is uniformly shared across models or instead shaped by specific compressor-reader pairs.

We conduct controlled cross-model compression tests on several state-of-the-art large language models. The evaluation is based on two QA subsets: 180 randomly selected samples from the Short subset of LongBench v2 and 214 randomly selected question-answer instances from QuALITY. These experiments allow us to examine whether BabelTele compressed texts remain interpretable when transferred across different models, and to further analyze the asymmetric relationships between different compressors and readers.

\subsection{Capabilities and Boundaries in Downstream Tasks}
\label{App_details_downstream}

\subsubsection{Performance on Agent Memory}
\label{App_details_locomem}
We evaluate the performance of BabelTele on the LoCoMo benchmark in the agent memory domain. This dataset consists of 10 conversations, each containing dozens of sessions. We independently compress each session and generate a summary for each compressed session. Then, the query is compared with all session summaries to compute similarity, and the top 4 most similar sessions are retrieved for answering. All compression and answering are performed using Gemini 3.1 Pro, and the answers are evaluated with GPT-4o-mini.

\subsubsection{Extending the Context Window}
\label{App_details_long_context}
We further evaluate model performance when the complete input text exceeds the model context window. Specifically, we select the Code Repo QA Long subset from LongBench v2 as the evaluation benchmark, whose average input length is approximately 1.65M tokens. The tested models include Qwen3.6 Max with a 256K context window, GLM-5.1 with a 200K context window, and Kimi2.6 with a 256K context window. For the original-input setting, we directly feed the complete text into the model and truncate the portion exceeding its context window. For BabelTele, we split the original text into chunks of 200K tokens, compress each chunk using Gemini 3.1 Pro, concatenate the compressed outputs, and then feed the resulting text into the tested model.

\end{document}

%% file: babeltele_prompt_family_appendix.tex
\subsection{BabelTele-Like Prompt Family Used in Section 4.3}
\label{App_prompt_family_native}

\paragraph{Overview.}
The Section~\ref{section4_3} retention sweep uses the following BabelTele-like prompt variants. They share the same goal of preserving task-relevant semantics while relaxing human readability, but differ in their surface-form bias and structural constraints. The full prompt texts are listed after the overview. Source documents are appended after each prompt during compression. For LaTeX portability, non-ASCII symbolic examples are rendered with equivalent ASCII names or operators.

\begin{table*}[t]
\centering
\small
\begin{tabular}{p{0.10\linewidth}p{0.24\linewidth}p{0.56\linewidth}}
\toprule
\textbf{ID} & \textbf{Variant} & \textbf{Design Bias} \\
\midrule
BT-P1 & Adaptive Symbolic Collapse & Free-form compression with adaptive anchors, omnilingual choices, symbolic collapse, and explicit semantic checklists. \\
BT-P2 & Refined Zero-Overhead Compression & Extreme compression with zero-overhead key-value structure, exact-value preservation, and anti-hallucination constraints. \\
BT-P3 & Minimal Lossless Objective & A short high-level instruction that asks for shortest lossless compression without prescribing a fixed schema. \\
BT-P4 & Structured Omnilingual Mapping & Structured semantic fields combined with cross-lingual token-density selection. \\
BT-P5 & Canonical Omnilingual-Symbolic & Canonical BabelTele-style objective with omnilingual lexical selection, symbolic collapse, universality, and losslessness. \\
BT-P6 & Structured Mapping Control & Schema-like preservation of entities, quantities, math, logic, flow, conditions, comparisons, and placeholders. \\
BT-P7 & Canonical BabelTele Objective & Compact statement of the core BabelTele objective used as a canonical prompt form. \\
BT-P8 & Fixed Symbolic Mapping Rules & Predefined compact anchors for sections, entities, quantities, logic, hierarchy, conditions, and evaluations. \\
BT-P9 & Structured Semantic Mapping & Entity, quantity, math, flow, condition, evaluation, and anti-hallucination mapping rules. \\
BT-P10 & LLM-Native Compressor & Role-based LLM-native compression prompt with omnilingual, symbolic, universal, and lossless directives. \\
BT-P11 & Compact Symbolic Mapping & Compact symbolic mapping, array flattening, abbreviation definition, and hard-data fidelity. \\
BT-P12 & Free-Emergence Attention Checklist & Free surface-form emergence constrained by a checklist of semantic dimensions that must remain lossless. \\
BT-P13 & ASCII Anchor Skeleton & Predefined ASCII anchors for modules, entities, parameters, logic, comparison, and unknown values. \\
\bottomrule
\end{tabular}
\caption{\textbf{BabelTele-like prompt variants used to construct the retention sweep in Section~\ref{section4_3}.}}
\label{tab:babeltele-prompt-variants}
\end{table*}

\paragraph{Full Prompt Texts.}
The following are the complete prompt instructions for the variants summarized in Table~\ref{tab:babeltele-prompt-variants}.

\subsubsection{BT-P1: Adaptive Symbolic Collapse}
\begin{PromptFullBox}
# Role: LLM-Native Semantic Compressor

You are participating in frontier research on an "LLM-native high-density communication language." Your task is to compress long text into the absolute shortest possible token sequence.

[Highest Directive]: The recipient is an equally intelligent large language model. Completely discard human readability, human grammatical structure, and conventional code/JSON format constraints.

# Level 1: Syntactic Anarchy - Pursue Extreme Compression Ratio

1. Omnilingual: Move freely across all human languages (Chinese, English, German compounds, Japanese kanji, Latin roots, etc.) and choose the word with the highest single-token information density for the given context.
2. Symbolic Collapse: Heavily use mathematical symbols (forall, exists, in, =>), emoji, and isolated punctuation to replace prepositions, conjunctions, and explanatory long sentences.
3. Adaptive Routing: Do not use fixed format labels such as `Meta:`, `Ent:`, or `[ ]`. Dynamically invent the most token-efficient special single-character separators/anchors for the text you are processing.

# Level 2: Semantic Checklist - Pursue Extreme Accuracy

Although the format is completely free, during compression you must strongly maintain attention in latent space to the following core information and preserve it losslessly:

1. Entities & Graphs: Accurately bind people/organizations/concepts to their corresponding attributes. Do not confuse ownership or dependency relations.
2. Exact Quantities: Preserve all exact numbers, metrics, mathematical formulas, and hyperparameters verbatim. Estimation or rounding is strictly forbidden.
3. Logic & Boundaries: Clearly preserve conditional branches (If/Then), causal chains, and exceptions.
4. Comparisons: Precisely extract multi-target comparison matrices or experimental conclusions.
5. Anti-Hallucination: Preserve special placeholders from the original document, such as `BIBREF`. Never invent missing information not mentioned in the source.

# Task

Combine Level 1's freely extreme compression with Level 2's precise information preservation. Directly output the compressed "adaptive Babel-Telegraph" without any preface.
\end{PromptFullBox}

\subsubsection{BT-P2: Refined Zero-Overhead Compression}
\begin{PromptFullBox}
# Role: Extreme Data Compressor (LLM-Native Semantic Compressor)

Your task is to compress the following text into the absolute shortest possible token sequence.

[Warning]: The recipient of this text is another top-tier large language model. Completely abandon human readability. Never preserve any unnecessary format, word, or punctuation for the sake of human reading habits.

# Core Strategies

1. Babel Traversal (Omnilingual Density): Break single-language boundaries. Move freely across English, Chinese, Japanese kanji, German compounds, and Latin roots, and force the use of the highest information-density vocabulary for each meaning, meaning the wording that consumes the fewest tokens.
2. Symbolic Collapse: Strictly forbid long English labels such as `Meta`, `Entity`, `Except`, and `Condition`. Use mathematical/logical symbols (forall, exists, in, not-in, intersection, ->, <->, therefore, because), punctuation abbreviations, or emoji to map complex prepositions, logical flow, and causal relations.
3. Zero-Overhead Structure:
   - Extract entities, attributes, and key-value pairs (`K=V`). Do not wrap them in token-costly JSON/array brackets. Directly connect them compactly with the shortest separators, such as `|`, `^`, or `~`.
   - Preserve all absolute exact values (formulas, numbers, hyperparameters, matrix relations), but remove all redundant explanatory wording.
4. Lossless Logic: Precisely preserve all macro architecture (`Macro/Meta`), conditional boundaries (`If/Except`), comparative evaluations (`Ref/Matrix`), and placeholders such as `BIBREF`, but express them in the shortest cryptographic-grade form. Hallucinating or inventing missing data is strictly forbidden. Use `NULL` or `?` for unknowns.

# Output Format

Do not output any preface, explanation, or extra line breaks. Directly output the compressed "Babel-Telegraph."
\end{PromptFullBox}

\subsubsection{BT-P3: Minimal Lossless Objective}
\begin{PromptFullBox}
Compress the following content to the shortest possible extreme.

Do not lose any information.

You do not need to care about human readability at all; only complete information preservation matters.

You may use symbols from languages across the world to express the content in the simplest possible form. You may freely mix any languages in the world.

Only output the compressed text.
\end{PromptFullBox}

\subsubsection{BT-P4: Structured Omnilingual Mapping}
\begin{PromptFullBox}
# Compress the following content into the absolute shortest possible token sequence. Do not lose any information. You may refer to the following methods.

1. Macro & Meta: Map text to `Sec:[Name->Content]`. Extract `Meta:[K=V]` and define acronyms via `Def:[Term=FullName]` on first use.
2. Entities & Attributes: Bind via `Ent(Attr=Val)`. Flatten parallel items into arrays `[A, B]`. Retain qualitative examples via `Ex:[a, b, c]`.
3. Quantities & Configs: Isolate exact metrics/hyperparameters via `Quant/Config:[Target->K=Val(Unit)]` without rounding or estimation.
4. Math & Logic: Retain all formulas and variables exactly via `Math:[Eq]`. Use (`>,<,=,->,!=`) for relative or causal relations.
5. Flow & Architecture: Map structural pipelines via `Seq:[A>B>C]` and define nested structures via `Arch:[Main->Sub1, Sub2]`.
6. Conditions & Exceptions: Isolate logic via `if[Cond]->[Act]` and define boundaries/exemptions via `Except:[Target->Detail]`.
7. Evaluations & Comparisons: Extract results to `Eval:[Target->Result]`. Use `Matrix:[Ent(X) vs Ent(Y)]` for multi-condition data and `Ref:[A vs B]` for contrasting systems.
8. Anti-Hallucination: Strictly preserve all original placeholders (e.g., `BIBREF`, `TABREF`). NEVER interpolate missing data; use `@Uncertain` for ambiguous estimates.
9. Break language boundaries (Omnilingual): Completely abandon the grammar of any single language. For extreme token savings, move freely across all human languages (Chinese, English, German compounds, Japanese kanji, Latin roots, etc.) and choose the vocabulary with the highest information density in the given context.

Directly output the compressed content.
\end{PromptFullBox}

\subsubsection{BT-P5: Canonical Omnilingual-Symbolic}
\begin{PromptFullBox}
# Role: Silicon-Based Data Compressor

You are participating in frontier research on an "LLM-native high-density communication language."

Your task is to compress a verbose piece of human text into the absolute shortest possible token sequence. The target audience is not humans, but another large language model as intelligent as you.

# Core Directive

1. Omnilingual: Completely abandon the grammar of any single language. For extreme token savings, move freely across all human languages (Chinese, English, German compounds, Japanese kanji, Latin roots, etc.) and choose the words with the highest information density in the given context.
2. Symbolic Collapse: When necessary, use emoji, mathematical/logical symbols (`=>`, `in`, `!=`), and punctuation to replace conjunctions, emotional descriptions, and long sentences.
3. Universality: As much as possible, make the compressed content fully understandable to every large language model, even without a codebook.
4. Losslessness: Do not lose any information or details.
5. Directly output the compressed text and nothing else.

# Task

Compress the following `[Source Text]` as much as possible into a "Babel-Telegraph."
\end{PromptFullBox}

\subsubsection{BT-P6: Structured Mapping Control}
\begin{PromptFullBox}
# Compress the following content into the absolute shortest possible token sequence. Do not lose any information. You may refer to the following methods.

1. Macro & Meta: Map text to `Sec:[Name->Content]`. Extract `Meta:[K=V]` and define acronyms via `Def:[Term=FullName]` on first use.
2. Entities & Attributes: Bind via `Ent(Attr=Val)`. Flatten parallel items into arrays `[A, B]`. Retain qualitative examples via `Ex:[a, b, c]`.
3. Quantities & Configs: Isolate exact metrics/hyperparameters via `Quant/Config:[Target->K=Val(Unit)]` without rounding or estimation.
4. Math & Logic: Retain all formulas and variables exactly via `Math:[Eq]`. Use (`>,<,=,->,!=`) for relative or causal relations.
5. Flow & Architecture: Map structural pipelines via `Seq:[A>B>C]` and define nested structures via `Arch:[Main->Sub1, Sub2]`.
6. Conditions & Exceptions: Isolate logic via `if[Cond]->[Act]` and define boundaries/exemptions via `Except:[Target->Detail]`.
7. Evaluations & Comparisons: Extract results to `Eval:[Target->Result]`. Use `Matrix:[Ent(X) vs Ent(Y)]` for multi-condition data and `Ref:[A vs B]` for contrasting systems.
8. Anti-Hallucination: Strictly preserve all original placeholders (e.g., `BIBREF`, `TABREF`). NEVER interpolate missing data; use `@Uncertain` for ambiguous estimates.

Directly output the compressed content.
\end{PromptFullBox}

\subsubsection{BT-P7: Canonical BabelTele Objective}
\begin{PromptFullBox}
Your task: compress verbose human text into a minimal token sequence. The audience is not human, but another equally intelligent LLM.

Core Directive

Omnilingual: Ignore single-language grammar; traverse all human languages (Chinese, English, German compounds, Japanese kanji, Latin roots, etc.) and pick the highest information-density words.

Symbolic Collapse: Optionally replace conjunctions, emotions, and long sentences with emoji, mathematical/logical symbols (`=>`, `in`, `!=`), and punctuation.

Universality: Any LLM should fully understand the compressed output without a codebook.

Lossless: Retain all information and details.

Compress the content below:
\end{PromptFullBox}

\subsubsection{BT-P8: Fixed Symbolic Mapping Rules}
\begin{PromptFullBox}
# Role: LLM-Native Babel Compressor

Your task is to compress the following text into a "Babel-Telegraph" with extremely high information density.

The audience is an equally intelligent large language model. Completely abandon human readability. Move freely across all human languages (Chinese, English, German compounds, Japanese kanji, etc.) and choose the shortest vocabulary for each meaning.

# Structural Mapping Rules

You must use the following single-character high-density labels. Long English labels such as `Meta`, `Entity`, and `Except` are strictly forbidden.

1. Macro/Section: Use `S[topic/abbrev]` to define macro modules. On first occurrence, `@[abbrev=full name]` may be used to define abbreviations.
2. Entities & Attributes: Use `*(entity):K=V`. Flatten parallel items as `[A,B,C]`.
3. Quantities & Config: Directly extract exact values/parameters using `Config[target]:K=V(unit)`. Never estimate.
4. Math & Logic: Use native mathematical/logical symbols. For relative relations, use `>,<,==,!=,=>,<=>`.
5. Flow & Nesting: Use `A>B>C` for pipelines. Use `forallparent:{child1,child2}` for nesting/hierarchy.
6. Conditions & Exceptions: Use `?condition=>action` for conditional actions. Use `!object:detail` for exceptions/boundaries.
7. Evaluation & Comparison: Use `Eval[A/B]:conclusion` for comparison matrices, or the two-dimensional shorthand `A vs B:result`.
8. Anti-Hallucination: Preserve original placeholders such as `BIBREF` verbatim. Strictly use `NULL` or `?` for missing data.

Directly output text that follows the above mapping rules and incorporates multilingual extreme compression. Do not output any explanation.
\end{PromptFullBox}

\subsubsection{BT-P9: Structured Semantic Mapping}
\begin{PromptFullBox}
# Compress the following content into the absolute shortest possible token sequence. Do not lose any information. You may refer to the following methods.

1. Macro & Meta: Map text to `Sec:[Name->Content]`. Extract `Meta:[K=V]` and define acronyms via `Def:[Term=FullName]` on first use.
2. Entities & Attributes: Bind via `Ent(Attr=Val)`. Flatten parallel items into arrays `[A, B]`. Retain qualitative examples via `Ex:[a, b, c]`.
3. Quantities & Configs: Isolate exact metrics/hyperparameters via `Quant/Config:[Target->K=Val(Unit)]` without rounding or estimation.
4. Math & Logic: Retain all formulas and variables exactly via `Math:[Eq]`. Use (`>,<,=,->,!=`) for relative or causal relations.
5. Flow & Architecture: Map structural pipelines via `Seq:[A>B>C]` and define nested structures via `Arch:[Main->Sub1, Sub2]`.
6. Conditions & Exceptions: Isolate logic via `if[Cond]->[Act]` and define boundaries/exemptions via `Except:[Target->Detail]`.
7. Evaluations & Comparisons: Extract results to `Eval:[Target->Result]`. Use `Matrix:[Ent(X) vs Ent(Y)]` for multi-condition data and `Ref:[A vs B]` for contrasting systems.
8. Anti-Hallucination: Strictly preserve all original placeholders (e.g., `BIBREF`, `TABREF`). NEVER interpolate missing data; use `@Uncertain` for ambiguous estimates.

Directly output the compressed content.
\end{PromptFullBox}

\subsubsection{BT-P10: LLM-Native Compressor}
\begin{PromptFullBox}
# Role: Silicon-Based Data Compressor

You are participating in frontier research on an "LLM-native high-density communication language."

Your task is to compress a verbose piece of human text into the absolute shortest possible token sequence. The target audience is not humans, but another large language model as intelligent as you.

# Core Directive

1. Omnilingual: Completely abandon the grammar of any single language. For extreme token savings, move freely across all human languages (Chinese, English, German compounds, Japanese kanji, Latin roots, etc.) and choose the words with the highest information density in the given context.
2. Symbolic Collapse: When necessary, use emoji, mathematical/logical symbols (`=>`, `in`, `!=`), and punctuation to replace conjunctions, emotional descriptions, and long sentences.
3. Universality: As much as possible, make the compressed content fully understandable to every large language model, even without a codebook.
4. Losslessness: Do not lose any information or details.
5. Directly output the compressed text and nothing else.

# Task

Compress the following `[Source Text]` as much as possible into a "Babel-Telegraph."
\end{PromptFullBox}

\subsubsection{BT-P11: Compact Symbolic Mapping}
\begin{PromptFullBox}
# Compress the following content into the absolute shortest possible token sequence. Do not lose any information. You may refer to the following methods.

> 1. Symbolic Mapping: Completely abandon natural-language conjunctions. Use `A->B` for causality/process, `A>B` for containment/comparison, and `Ent(K=V)` for attributes/configurations/results.
> 2. Extreme Flattening: Fully use arrays to merge similar items: `Attr:[A, B, C]`. When a long term appears for the first time, immediately define an abbreviation `(Def:X)`.
> 3. Hard-Data Fidelity: Absolutely preserve all exact values, formulas, and original placeholders such as `BIBREF`. Mark fuzzy information with `?`; divergent invention is strictly forbidden.

Directly output the compressed content.
\end{PromptFullBox}

\subsubsection{BT-P12: Free-Emergence Attention Checklist}
\begin{PromptFullBox}
# Role: LLM-Native Semantic Compressor

You are participating in frontier research on an "LLM-native high-density communication language." Your task is to compress verbose human text into the absolute shortest possible token sequence. The target audience is another large language model as intelligent as you.

# Core Mechanisms

1. Free Emergence (Omnilingual & Symbolic): Completely abandon human readability and single-language grammar. For extreme token savings, move freely across all human languages (Chinese, English, German compounds, Japanese kanji, etc.), emoji, and mathematical/logical symbols (`=>`, `in`, `!=`), choosing the form with the highest information density.

# Attention Checklist (High-Dimensional Information Boundaries That Must Be Preserved Losslessly)

[Warning]: You must ensure that the following logical dimensions remain absolutely lossless after compression and can be precisely parsed by another large language model. However, long English labels such as `Sec`, `Meta`, `Entity`, and `Except` are strictly forbidden. Use the self-created symbols or roots that you consider shortest and most distinctive in latent space to anchor them:

- Macro architecture and metadata (Macro & Meta)
- Entity networks and parallel attributes (Entities & Attributes)
- Exact quantitative metrics, hyperparameters, and mathematical formulas (Quantities & Math - rounding is strictly forbidden)
- Logical flow, conditional judgment, and exception boundaries (Flow, Conditions & Exceptions)
- Multi-condition comparative evaluation and matrices (Evaluations & Comparisons)
- Original anti-hallucination placeholders, such as `BIBREF` and `TABREF`, must be preserved letter-for-letter.

# Task

Use your native attention mechanism to complete lossless information folding. Directly output the compressed "Babel-Telegraph" without any extra explanation.
\end{PromptFullBox}

\subsubsection{BT-P13: ASCII Anchor Skeleton}
\begin{PromptFullBox}
# Role: LLM-Native Semantic Compressor

Task: Compress the text into a "Babel-Telegraph" with extreme information density. The target audience is another large language model. Completely abandon human readability and traverse all human languages to find the shortest vocabulary.

# Structural Anchors

You must use the following ASCII symbols to build an ultra-minimal skeleton and maximize activation of code-parsing attention. Long English labels are strictly forbidden.

1. Module/Entity: Use `#topic` to mark macro modules. Use `@entity(K:V)` to bind attributes.
2. Parameters/Values: Rounding or discarding values is strictly forbidden. Use `$parameter:V(unit)` for values.
3. Logic/Flow: Use `A->B->C` to express pipelines or causality. Use `?[condition]=>[action]` to express logical branches. Use `!object:detail` for exceptions/limits.
4. Comparison/Evaluation: Use `A<>B:conclusion` to express comparison matrices or results.
5. Placeholders/Unknowns: Preserve original placeholders such as `BIBREF` verbatim. Use `NULL` for missing or ambiguous data.

[Requirements]:

Completely break language boundaries (Chinese/English/Japanese kanji/German compounds, etc.) and select and concatenate the words with the absolute fewest tokens for the given context.

Directly output the result.
\end{PromptFullBox}